\documentclass{article}

\usepackage[preprint]{neurips_2023}

\usepackage[utf8]{inputenc} 
\usepackage[T1]{fontenc}    
\usepackage{hyperref}       
\usepackage{url}            
\usepackage{booktabs}       
\usepackage{amsfonts}       
\usepackage{nicefrac}       
\usepackage{microtype}      
\usepackage{xcolor}         

\usepackage{graphicx}
\usepackage{comment}
\usepackage{multirow}
\usepackage{amsmath}
\usepackage{tablefootnote}
\usepackage{subfigure}
\newcommand{\kho}[1]{\textcolor{black}{#1}}
\usepackage{algorithm}
\usepackage{algorithmic}

\title{Residual Scheduling: A New Reinforcement Learning Approach to Solving Job Shop Scheduling Problem}

\author{%
  Kuo-Hao Ho\\ 
  Department of Computer Science\\
  National Yang Ming Chiao Tung University\\
  No. 1001, Daxue Rd., Hsinchu City, Taiwan\\
  \texttt{lukewayne123.cs05@nycu.edu.tw} \\
  \And
  Ruei-Yu Jheng\\
  Department of Computer Science\\
  National Yang Ming Chiao Tung University\\
  No. 1001, Daxue Rd., Hsinchu City, Taiwan\\
  \texttt{rayy.11@nycu.edu.tw} \\
  \And
  Ji-Han Wu\\
  Department of Computer Science\\
  National Yang Ming Chiao Tung University\\
  No. 1001, Daxue Rd., Hsinchu City, Taiwan\\
  \texttt{oo12374@gmail.com} \\
  \And
  Fan Chiang\\
  Department of Computer Science\\
  National Yang Ming Chiao Tung University\\
  No. 1001, Daxue Rd., Hsinchu City, Taiwan\\
  \texttt{lalwin6404@gmail.com} \\
  \And
  Yen-Chi Chen\\
  Research Center for Information Technology Innovation\\
  Academia Sinica\\
  128 Academia Road, Taipei, Taiwan\\
  \texttt{zxkyjimmy@gmail.com} \\
  \And
  Yuan-Yu Wu\\
  Department of Computer Science\\
  National Yang Ming Chiao Tung University\\
  No. 1001, Daxue Rd., Hsinchu City, Taiwan\\
  \texttt{warren03wu@gmail.com} \\
  \And
  I-Chen Wu\\
  Department of Computer Science / Research Center for Information Technology Innovation\\
  National Yang Ming Chiao Tung University / Academia Sinica\\
  No. 1001, Daxue Rd., Hsinchu City, Taiwan / 128 Academia Road, Taipei, Taiwan\\
  \texttt{icwu@cs.nctu.edu.tw} \\
}

\begin{document}

\maketitle

\begin{abstract}
Job-shop scheduling problem (JSP) is a mathematical optimization problem widely used in industries like manufacturing, and flexible JSP (FJSP) is also a common variant. 
Since they are NP-hard, it is intractable to find the optimal solution for all cases within reasonable times.
Thus, it becomes important to develop efficient heuristics to solve JSP/FJSP.
A kind of method of solving scheduling problems is construction heuristics, which constructs scheduling solutions via heuristics.
Recently, many methods for construction heuristics leverage deep reinforcement learning (DRL) with graph neural networks (GNN).
In this paper, we propose a new approach, named residual scheduling, to solving JSP/FJSP.
In this new approach, we remove irrelevant machines and jobs such as those finished, such that the states include the remaining (or relevant) machines and jobs only. 
Our experiments show that our approach reaches state-of-the-art (SOTA) among all known construction heuristics on most well-known open JSP and FJSP benchmarks. 
In addition, we also observe that even though our model is trained for scheduling problems of smaller sizes, our method still performs well for scheduling problems of large sizes. 
Interestingly in our experiments, our approach even reaches \kho{zero makespan gap} for 49 among 50 JSP instances whose job numbers are more than 150 on 20 machines. 

\end{abstract}

\section{Introduction}
The \textit{job-shop scheduling problem} (\textit{JSP}) is a \kho{combinatorial optimization} (CO) problem widely used in many industries, like manufacturing \citep{L2D_NIPS_ZhangSC0TX20,Waschneck2016ProductionSI}. 
For example, a semiconductor manufacturing process can be viewed as a complex JSP problem \citep{Waschneck2016ProductionSI}, where a set of given jobs are assigned to a set of machines under some constraints to achieve some expected goals such as minimizing makespan which is focused on in this paper. 
While there are many variants of JSP \citep{JSPtop-downSurveyAbdolrazzaghNezhad2017JobSS}, we also consider an extension called \textit{flexible JSP} (\textit{FJSP}) where job operations can be done on designated machines. 

A generic approach to solving \kho{CO} problems is to use mathematical programming, such as mixed integer linear programming (MILP)\kho{, or constraint programming (CP)}. 
Two popular generic \kho{CO} solvers for solving \kho{CO} are \textit{OR-Tools} \citep{ortools} and \textit{IBM ILOG CPLEX Optimizer} (abbr. \textit{CPLEX}) \citep{cplex2009v12}. 
However, both JSP and FJSP, as well as many other \kho{CO} problems, have been shown to be NP-hard \citep{NP-hard_Garey1978ComputersAI,NP-hard_Lageweg1977JobShopSB}.
That said, it is unrealistic and intractable to find the optimal solution for all cases within reasonable times. 
These tools can obtain the optimal solutions if sufficient time (or unlimited time) is given; otherwise, return best-effort solutions during the limited time, which usually have gaps to the optimum.
When problems are scaled up, the gaps usually grow significantly. 

In practice, some heuristics \citep{heuristicRules_Gupta2006JobSS,PDRdef_heuristicRules_Haupt1989ASO} or approximate methods \citep{approximateMethod_Jansen2000ApproximationAF} were used to cope with the issue of intractability. 
A simple greedy approach is to use the heuristics following the so-called \textit{priority dispatching rule} (\textit{PDR}) \citep{PDRdef_heuristicRules_Haupt1989ASO} to construct solutions. 
These can also be viewed as a kind of \textit{solution construction heuristics} or \textit{construction heuristics}.
Some of PDR examples are \emph{First In First Out} (\emph{FIFO}), \emph{Shortest Processing Time} (\emph{SPT}), \emph{Most WorK Remaining} (\emph{MWKR}), and \emph{Most Operation Remaining} (\emph{MOR}).
Although these heuristics are usually computationally fast, it is hard to design generally effective rules to minimize the gap to the optimum, and the derived results are usually far from the optimum. 

Furthermore, a generic approach to \kho{ search within a search space of problem solutions} is called \emph{metaheuristics}, such as tabu search 
, genetic algorithm (GA) \citep{GA_Pezzella2008AGA,costGA_Ren2012ANH},
and PSO algorithms \citep{PSO_journals/amc/LianJG06,PSO_journals/eswa/LiuSYK11}. 
However, metaheuristics still take a high computation time, and it is not ensured to obtain the optimal solution either. 

Recently, deep reinforcement learning (DRL) has made several significant successes for some applications, such as AlphaGo \citep{AlphaGo_Silver2016MasteringTG}, AlphaStar \citep{AlphaStar_Vinyals2019GrandmasterLI}, AlphaTensor \citep{AlphaTensor_fawzi2022discovering}, and thus it also attracted much attention in the \kho{CO} problems, including chip design \citep{Chip_Mirhoseini2021AGP} and scheduling problems \citep{Review_JSPRL_2023}.
In the past, several researchers used DRL methods as construction heuristics, and their methods did improve scheduling performance, illustrated as follows. 
\citet{KoreaRLJSP_Park2020ARL} proposed a method based on DQN \citep{DQN_nature/MnihKSRVBGRFOPB15} for JSP in semiconductor manufacturing and showed that their DQN model outperformed GA in terms of both scheduling performance (namely gap to the optimum on makespan) and computation time.
\citet{PDRrules_LinDCC19} and \citet{BeijingNTHU_Luo2020DynamicSF} proposed different DQN models to decide the scheduling action among the heuristic rules and improved the makespan and the tardiness over PDRs, respectively.

A recent DRL-based approach to solving JSP/FJSP problems is to leverage graph neural networks (GNN) to design a size-agnostic representation \citep{L2D_NIPS_ZhangSC0TX20,GNNDRL_Park2021LearningTS,ScheduleNet_abs-2106-03051,FJSP_L2D_SongCLC23}.
In this approach, graph representation has better generalization ability in larger instances and provides a holistic view of scheduling states.
\citet{L2D_NIPS_ZhangSC0TX20} proposed a DRL method with disjunctive graph representation for JSP, called \textit{L2D} (\textit{Learning to Dispatch}), and used GNN to encode the graph for scheduling decision.
Besides, \citet{FJSP_L2D_SongCLC23} extended their methods to FJSP.
\citet{GNNDRL_Park2021LearningTS} used a similar strategy of \citep{L2D_NIPS_ZhangSC0TX20} but with different state features and model structure. 
\citet{ScheduleNet_abs-2106-03051} proposed a new approach to solving JSP, called \textit{ScheduleNet}, by using a different graph representation and a DRL model with the graph attention for scheduling decision.
Most of the experiments above showed that their models trained from small instances still worked reasonably well for large test instances, and generally better than PDRs. 
Among these methods, ScheduleNet achieved state-of-the-art (SOTA) performance.
There are still other DRL-based approaches to solving JSP/FJSP problems, but not construction heuristics.
\citet{L2S_2022} proposes another approach, called Learning to Search (L2S), a kind of search-based heuristics. 
In this paper, we propose a new approach to solving JSP/FJSP, a kind of construction heuristics, also based on GNN. 
In this new approach, we remove irrelevant machines and jobs, such as those finished, such that the states include the remaining machines and jobs only. 
This approach is named \textit{residual scheduling} in this paper to indicate to work on the remaining graph. 

Without irrelevant information, our experiments show that our approach reaches SOTA by outperforming the above mentioned construction methods on some well-known open benchmarks, seven for JSP and two for FJSP, as described in Section \ref{sec:experiments}. 
We also observe that even though our model is trained for scheduling problems of smaller sizes, our method still performs well for scheduling problems of large sizes. 
Interestingly in our experiments, our approach even reaches \kho{zero makespan gap} for 49 among 50 JSP instances whose job numbers are more than 150 on 20 machines. 

\section{Problem Formulation}
\label{sec:formulation}

\subsection{JSP and FJSP}
\label{sec:jsp}
\begin{figure}[h]
    \centering
    \subfigure[]{
    \includegraphics[width=0.23\textwidth]{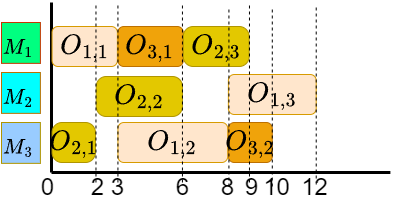}
    \label{fig:JSP_opt_solution}
    }
    \subfigure[]{
    \includegraphics[width=0.23\textwidth]{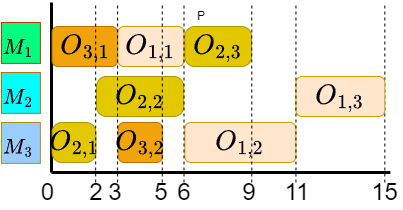}
    \label{fig:JSP_no_opt_solution}
    }
    \subfigure[]{
    \includegraphics[width=0.23\textwidth]{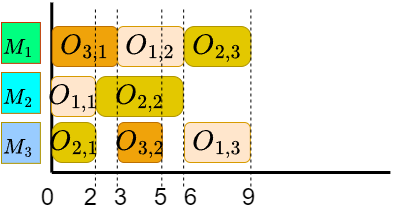}
    \label{fig:FJSP_opt_solution}
    }
    \subfigure[]{
    \includegraphics[width=0.23\textwidth]{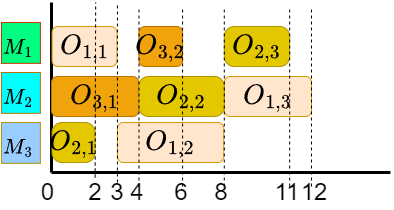}
    \label{fig:FJSP_no_opt_solution}
    }
    \caption{Both (a) and (b) are solutions of the 3x3 JSP instance in Table \ref{tab:3x3JSPexample} (a), and the former has the minimal makespan, 12. Both (c) and (d) are solutions of the 3x3 FJSP instance in Table \ref{tab:3x3JSPexample} (b), and the former has the minimal makespan, 9.}
    \label{fig:solutions}
\end{figure}

A $n \times m$ JSP instance contains $n$ jobs and $m$ machines. 
Each job $J_j$ consists of a sequence of $k_j$ operations $\{O_{j,1},\dots, O_{j,k_j}\}$, where operation $O_{j,i}$ must be started after $O_{j,i-1}$ is finished.
One machine can process at most one operation at a time, and preemption is not allowed upon processing operations.
In JSP, one operation $O_{j,i}$ is allowed to be processed on one designated machine, denoted by $M_{j,i}$, with a processing time, denoted by $T^{(op)}_{j,i}$. 
Table \ref{tab:3x3JSPexample} (a) illustrates a $3 \times 3$ JSP instance, where the three jobs have 3, 3, 2 operations respectively, each of which is designated to be processed on one of the three machines $\{M_1, M_2, M_3\}$ in the table.
A solution of a JSP instance is to dispatch all operations $O_{j,i}$ to the corresponding machine $M_{j,i}$ at time $\tau^{(s)}_{j,i}$, such that the above constraints are satisfied. 
Two solutions of the above 3x3 JSP instance are given in Figure \ref{fig:solutions} (a) and (b). 


\begin{table}[h]
\centering
\caption{JSP and FJSP instances}
\label{tab:3x3JSPexample}
    \begin{subtable}[A $3 \times 3$ JSP instance]{
    \label{tab:3x3JSP_instance}
    \begin{tabular}{|c|c|c|c|c|}
    \hline
        Job & Operation & $M_1$ & $M_2$ & $M_3$\\
        \hline
        \multirow{3}{*}{Job 1} & $O_{1,1}$ & 3 &  & \\
        \cline{2-5}
         & $O_{1,2}$ & & & 5\\
        \cline{2-5}
         & $O_{1,3}$ & & 4 & \\
        \hline
        \multirow{3}{*} {Job 2} & $O_{2,1}$ & & & 2\\
        \cline{2-5}
         & $O_{2,2}$ & & 4 & \\
         \cline{2-5}
         & $O_{2,3}$ & 3 & & \\
        \hline
        \multirow{2}{*} {Job 3} & $O_{3,1}$ & 3 & & \\
        \cline{2-5}
         & $O_{3,2}$ &  & & 2\\
    \hline
    \end{tabular}
    }
    \end{subtable}
    \begin{subtable}[A $3 \times 3$ FJSP instance]{
    \label{tab:3x3FJSP_instance}
    \begin{tabular}{|c|c|c|c|c|}
    \hline
        Job & Operation & $M_1$ & $M_2$ & $M_3$\\
        \hline
        \multirow{3}{*}{Job 1} & $O_{1,1}$ & 3 & 2 & \\
        \cline{2-5}
         & $O_{1,2}$ & 3 & & 5\\
        \cline{2-5}
         & $O_{1,3}$ & & 4 & 3 \\
        \hline
        \multirow{3}{*} {Job 2} & $O_{2,1}$ & & & 2\\
        \cline{2-5}
         & $O_{2,2}$ & & 4 & \\
         \cline{2-5}
         & $O_{2,3}$ & 3 & & \\
        \hline
        \multirow{2}{*} {Job 3} & $O_{3,1}$ & 3 & 4 & \\
        \cline{2-5}
         & $O_{3,2}$ & 2 & & 2\\
    \hline
    \end{tabular}
    }
    \end{subtable}
    
\end{table}

While there are different expected goals, such as makespan, tardiness, etc., this paper focuses on makespan. 
Let the first operation start at time $\tau=0$ in a JSP solution initially. 
The makespan of the solution is defined to be $T^{(mksp)} = \max(\tau^{(c)}_{j,i})$ for all operations $O_{j,i}$, where $\tau^{(c)}_{j,i}=\tau^{(s)}_{j,i} + T^{(op)}_{j,i}$ denotes the completion time of $O_{j,i}$. 
The makespans for the two solutions illustrated in Figure \ref{fig:solutions} (a) and (b) are 12 and 15 respectively. 
The objective is to derive a solution that minimizes the makespan $T^{(mksp)}$, and  
the solution of Figure \ref{fig:solutions} (a) reaches the optimal.  

A $n \times m$ FJSP instance is also a $n \times m$ JSP instance with the following difference. 
In FJSP, all operations $O_{j,i}$ are allowed to be dispatched to multiple designated machines with designated processing times. 
Table \ref{tab:3x3JSPexample} (b) illustrates a $3 \times 3$ FJSP instance, where multiple machines can be designated to be processed for one operation. 
Figure \ref{fig:solutions} (c) illustrates a solution of an FJSP instance, which takes a shorter time than that in Figure \ref{fig:solutions} (d).

\subsection{Construction Heuristics}
\label{sec:constuction}

An approach to solving these scheduling problems is to construct solutions step by step in a greedy manner, and the heuristics based on this approach is called \textit{construction heuristics} in this paper. 
In the approach of construction heuristics, a scheduling solution is constructed through a sequence of partial solutions in a chronicle order of dispatching operations step by step, defined as follows. 
The $t$-th partial solution $S_{t}$ associates with a \textit{dispatching time} $\tau_{t}$ and includes a partial set of operations that have been dispatched by $\tau_{t}$ (inclusive) while satisfying the above JSP constraints, and all the remaining operations must be dispatched after $\tau_{t}$ (inclusive). 
The whole construction starts with $S_0$ where none of operations have been dispatched and the dispatching time is $\tau_0=0$. 
For each $S_t$, a set of operations to be chosen for dispatching form a set of pairs of ($M$, $O$), called \textit{candidates} $C_t$, where operations $O$ are allowed to be dispatched on machines $M$ at $\tau_{t}$.
An agent (or a heuristic algorithm) chooses one from candidates $C_t$ for dispatching, and transits the partial solution to the next $S_{t+1}$. 
If there exists no operations for dispatching, the whole solution construction process is done and the partial solution is a solution, since no further operations are to be dispatched. 

\begin{figure}[h]
    \centering
    \subfigure[$S_0$]{
    \includegraphics[width=0.3\textwidth,height=1.8cm]{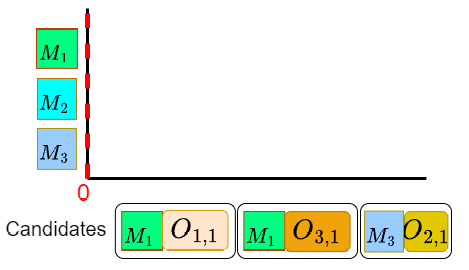}
    \label{fig:JSP_solution_0}
    }
    \subfigure[$S_1$]{
    \includegraphics[width=0.3\textwidth,height=1.8cm]{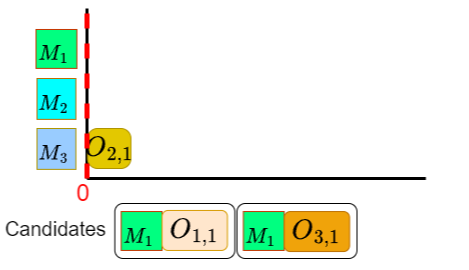}
    \label{fig:JSP_solution_1}
    }
    \subfigure[$S_2$]{
    \includegraphics[width=0.3\textwidth,height=1.8cm]{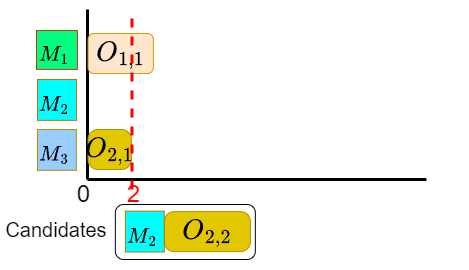}
    \label{fig:JSP_solution_2}
    }
    \subfigure[$S_3$]{
    \includegraphics[width=0.3\textwidth,height=1.8cm]{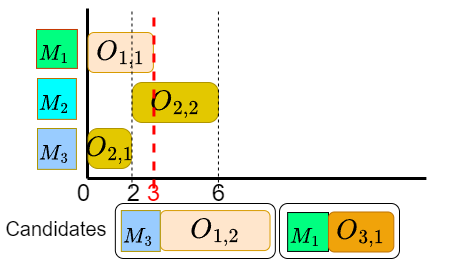}
    \label{fig:JSP_solution_3}
    }
    \subfigure[$S_4$]{
    \includegraphics[width=0.3\textwidth,height=1.8cm]{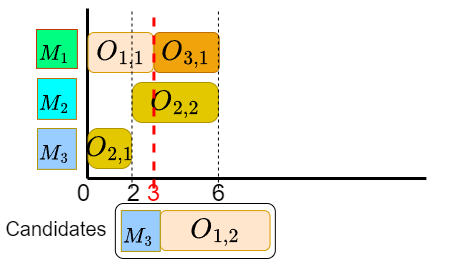}
    \label{fig:JSP_solution_4}
    }
    \subfigure[$S_5$]{
    \includegraphics[width=0.3\textwidth,height=1.8cm]{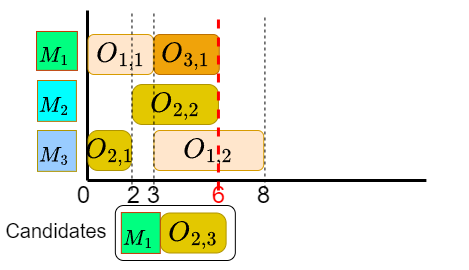}
    \label{fig:JSP_solution_5}
    }
    \subfigure[$S_6$]{
    \includegraphics[width=0.3\textwidth,height=1.8cm]{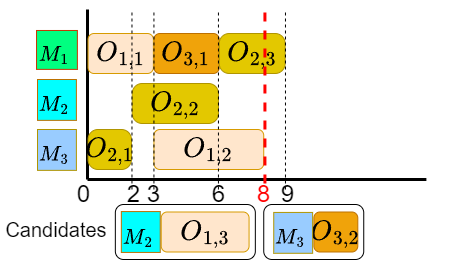}
    \label{fig:JSP_solution_6}
    }
    \subfigure[$S_7$]{
    \includegraphics[width=0.3\textwidth,height=1.8cm]{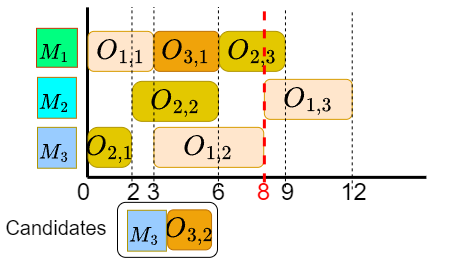}
    \label{fig:JSP_solution_7}
    }
    \subfigure[$S_8$]{
    \includegraphics[width=0.3\textwidth,height=1.8cm]{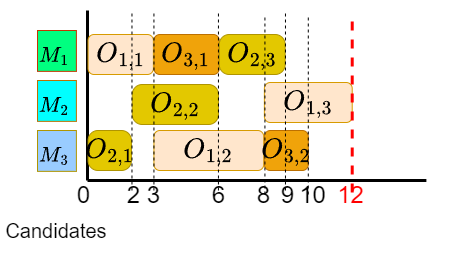}
    \label{fig:JSP_solution_8}
    }
    \caption{Solution construction, a sequence of partial solutions from $S_0$ to $S_8$.}
    \label{fig:partial_solution_steps}
\end{figure}

Figure \ref{fig:partial_solution_steps} illustrates a solution construction process for the 3x3 JSP instance in Table \ref{tab:3x3JSP_instance}, constructed through nine partial solutions step by step. 
The initial partial solution $S_0$ starts without any operations dispatched as in Figure \ref{fig:partial_solution_steps} (a). 
The initial candidates $C_0$ are $\{(M_1, O_{1,1}),(M_3, O_{2,1}),(M_1, O_{3,1})\}$. 
Following some heuristic, construct a solution from partial solution $S_0$ to $S_9$ step by step as in the Figure, where the dashed line in red indicate the time $\tau_t$. 
The last one $S_9$, the same as the one in Figure \ref{fig:solutions} (a), is a solution, since all operations have been dispatched, and the last operation ends at time 12, the makespan of the solution. 

For FJSP, the process of solution construction is almost the same except for that one operation have multiple choices from candidates.  
Besides, an approach based on solution construction can be also viewed as the so-called \textit{Markov decision process} (\textit{MDP}), and the MDP formulation for solution construction is described in more detail in the appendix.

\section{Our Approach}
\label{sec:approach}

In this section, we present a new approach, called \textit{residual scheduling}, to solving scheduling problems.
We introduce the residual scheduling in Subsection \ref{sec:revisit-partial-sol}, 
describe the design of the graph representation in Subsection \ref{sec:residual-graph-rep}, 
propose a model architecture based on graph neural network in Subsection \ref{sec:gnn} and 
present a method to train this model in Subsection \ref{sec:training};

\subsection{Residual Scheduling}
\label{sec:revisit-partial-sol}

In our approach, the key is to remove irrelevant information, particularly for operations, from states (including partial solutions).  
An important benefit from this is that we do not need to include all irrelevant information while training to minimize the makespan.
Let us illustrate by the state for the partial solution $S_3$ at time $\tau_3=3$ in Figure \ref{fig:partial_solution_steps} (d). 
All processing by $\tau_3$ are irrelevant to the remaining scheduling. 
Since operations $O_{1,1}$ and $O_{2,1}$ are both finished and irrelevant the rest of scheduling, they can be removed from the state of $S_3$. 
In addition, operation $O_{2,2}$ is dispatched at time 2 (before $\tau_3=3$) and its processing time is $T^{(op)}_{2,1}=4$, so the operation is marked as \textit{ongoing}. 
Thus, the operation can be modified to start at $\tau_3=3$ with a processing time $4-(3-2)$. 
Thus, the modified state for $S_3$ do not contain both $O_{1,1}$ and $O_{2,1}$, and modify $O_{2,2}$ as above. 
Let us consider two more examples. 
For $S_4$, one more operation $O_{2,2}$ is dispatched and thus marked as ongoing, however, the time $\tau_4$ remains unchanged and no more operations are removed.
In this case, the state is almost the same except for including one more ongoing operation $O_{2,2}$.
Then, for $S_5$, two more operations $O_{3,1}$ and $O_{2,2}$ are removed and the ongoing operation $O_{1,2}$ changes its processing time to the remaining time (5-3). 

For residual scheduling, we also reset the dispatching time $\tau=0$ for all states with partial solutions modified as above, so we derive makespans which is also irrelevant to the earlier operations.  
Given a scheduling policy $\pi$, $T^{(mksp)}_{\pi}(S)$ is defined to be the makespan derived from an episode starting from states $S$ by following $\pi$, and $T^{(mksp)}_{\pi}(S,a)$ the makespan by taking action $a$ on $S$. 

\subsection{Residual Graph Representation}
\label{sec:residual-graph-rep}

\begin{figure}[h]
    \centering
    \subfigure[Graph for $S_3$]{
    \includegraphics[width=0.25\textwidth]{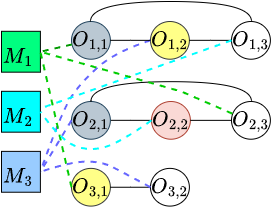}
    \label{fig:JSP_solution_graph_3}
    }
    \subfigure[Graph embedding]{
    \includegraphics[width=0.4\textwidth]{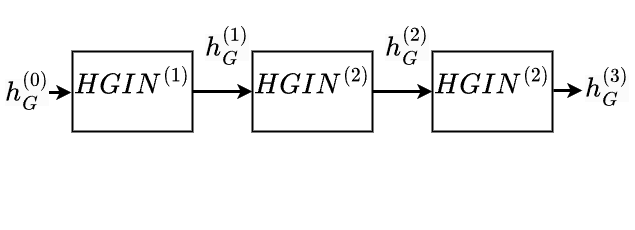}
    \label{fig:graph_embedding_phase}
    }
    \subfigure[Score function]{
    \includegraphics[width=0.3\textwidth]{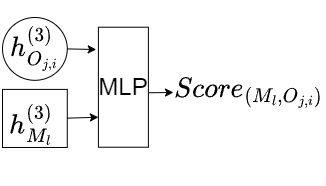}
    \label{fig:score_function}
    }
    \caption{Graph representation and networks.}
    \label{fig:networks_for_our_approach}
\end{figure}

In this paper, our model design is based on graph neural network (GNN), and leverage GNN to extract the scheduling decision from the relationship in graph.
In this subsection, we present the graph representation. 
Like many other researchers such as \citet{ScheduleNet_abs-2106-03051}, we formulate a partial solution into a graph $\mathcal{G}=(\mathcal{V}, \mathcal{E})$, where $\mathcal{V}$ is a set of nodes and $\mathcal{E}$ is a set of edges.
A node is either a machine node $M$ or an operation node $O$.
An edge connects two nodes to represent the relationship between two nodes, basically including three kinds of edges, namely operation-to-operation ($O \to O$), machine-to-operation ($M \to O$) and operation-to-machine ($O \to M$).
All operations in the same job are fully connected as $O \to O$ edges.
If an operation $O$ is able to be performed on a machine $M$, there exists both  $O \to M$ and $M \to O$ directed edges.
In \citep{ScheduleNet_abs-2106-03051}, they also let all machines be fully connected as $M \to M$ edges. 
However, our experiments in section \ref{sec:experiments} show that mutual $M \to M$ edges do not help much based on our Residual Scheduling. 
An illustration for graph representation of $S_3$ is depicted in Figure \ref{fig:networks_for_our_approach} (a). 

In the graph representation, all nodes need to include some attributes so that a partial solution $S$ at the dispatching time $\tau$ can be supported in the MDP formulation (in the appendix). 
Note that many of the attributes below are normalized to reduce variance. 
For nodes corresponding to operations $O_{j,i}$, we have the following attributes:

\emph{\textbf{Status}} $\phi_{j,i}$:
        The operation status $\phi_{j,i}$ is 
        \emph{completed} if the operation has been finished by $\tau$, 
        \emph{ongoing} if the operation is ongoing (i.e., has been dispatched to some machine by $\tau$ and is still being processed at $\tau$), 
        \emph{ready} if the operation designated to the machine which is idle has not been dispatched yet and its precedent operation has been finished, and
        \emph{unready} otherwise. 
        For example, in Figure \ref{fig:networks_for_our_approach} (a), the gray nodes are \emph{completed}, the red \emph{ongoing}, the yellow \emph{ready} and the white \emph{unready}. 
        In our residual scheduling, there exists no completed operations in all partial solutions, since they are removes for irrelevance of the rest of scheduling. 
        The attribute is a one-hot vector to represent the current status of the operation, which is one of \emph{ongoing}, \emph{ready} and \emph{unready}.
        Illustration for all states $S_0$ to $S_8$ are shown in the appendix.

\emph{\textbf{Normalized processing time}} $\bar{T}^{(op)}_{j,i}$: 
        Let the maximal processing time be $T^{(op)}_{max}=\max_{\forall j,i}(T^{(op)}_{j,i})$.
        Then,  $\bar{T}^{(op)}_{j,i}=T^{(op)}_{j,i}/T^{(op)}_{max}$. 
        In our residual scheduling, the operations that have been finished are removed in partial solutions and therefore their processing time can be ignored; 
        the operations that has not been dispatched yet still keep their processing times the same;
        the operations that are \textit{ongoing} change their processing times to the remaining times after the dispatching time $\tau_{t}$. 
        As for FJSP, the operations that has not been dispatched yet may have several processing times on different machines, and thus we can simply choose the average of these processing times.

\emph{\textbf{Normalized job remaining time}} $\bar{T}^{(job)}_{j,i}$: 
        Let the rest of processing time for job $J_j$ be $T^{(job)}_{j,i}=\sum_{\forall i' \geq i}T^{(op)}_{j,i'}$, and let the processing time for the whole job $j$ be $T^{(job)}_{j}=\sum_{\forall i'}T^{(op)}_{j,i'}$. 
        In practice, $T^{(job)}_{j}$ is replaced by the processing time for the original job $j$.
        Thus, $\bar{T}^{(job)}_{j,i} = T^{(job)}_{j,i}/T^{(job)}_{j}$. 
        For FJSP, since operations $O_{j,i}$ can be dispatched to different designated machines $M_l$, say with the processing time $T^{(op)}_{j,i,l}$, we simply let $T^{(op)}_{j,i}$ be the average of $T^{(op)}_{j,i,l}$ for all $M_l$. 

For machine nodes corresponding to machines $M_l$, we have the following attributes: 

\emph{\textbf{Machine status}} $\phi_l$:
        The machine status $\phi_{l}$ is 
        \emph{processing} if some operation has been dispatched to and is being processed by $M_l$ at $\tau$, and
        \emph{idle} otherwise (no operation is being processed at $\tau$).  
        The attribute is a one-hot vector to represent the current status, which is one of \emph{processing} and \emph{idle}.

\emph{\textbf{Normalized operation processing time}} $\bar{T}^{(mac)}_l$:
        On the machine $M_l$, the processing time $T^{(mac)}_l$ is $T^{(op)}_{j,i}$ (the same as the normalized processing time for node $O_{j,i}$) if the machine status is \textit{processing}, i.e., some ongoing operation $O_{j,i}$ is being processed but not finished yet, is zero if the machine status is \textit{idle}. 
        Then, this attribute is normalized to $T^{(op)}_{max}$ and thus $\bar{T}^{(mac)}_l = T^{(mac)}_l / T^{(op)}_{max}$.

Now, consider edges in a residual scheduling graph. 
As described above, there exists three relationship sets for edges, $O \to O$, $O \to M$ and $M \to O$. 
First, for the same job, say $J_j$, all of its operation nodes for $O_{j,i}$ are fully connected. 
Note that for residual scheduling the operations finished by the dispatching time $\tau$ are removed and thus have no edges to them. 
Second, a machine node for $M_l$ is connected to an operation node for $O_{j,i}$, if the operation $O_{j,i}$ is designated to be processed on the machine $M_l$, which forms two edges $O \to M$ and $M \to O$.
Both contains the following attribute. 

\emph{\textbf{Normalized operation processing time}} $\bar{T}^{(edge)}_{j,i,l}$:
        The attribute is $\bar{T}^{(edge)}_{j,i,l} = T^{(op)}_{j,i}/T^{(op)}_{max}$. 
        Here, $T^{(op)}_{j,i} = T^{(op)}_{j,i,l}$ in the case of FJSP. 
        If operation $O_{j,i}$ is ongoing (or being processed), $T^{(op)}_{j,i}$ is the remaining time as described above.

\subsection{Graph Neural Network}
\label{sec:gnn}

In this subsection, we present our model based on graph neural network (GNN).
GNN are a family of deep neural networks \citep{GNN_Battaglia2018RelationalIB} that can learn representation of graph-structured data, widely used in many applications \citep{GNN_Lv2021TemporalMC,GNN_Zhou2020GraphNN}.
A GNN aggregates information from node itself and its neighboring nodes and then update the data itself, which allows the GNN to capture the complex relationships within the data graph. 
For GNN, we choose \textit{Graph Isomorphism Network} (\textit{GIN}), which was shown to have strong discriminative power \citep{GIN_Xu2019HowPA} and summarily reviewed as follows.
Given a graph $\mathcal{G} = (\mathcal{V}, \mathcal{E})$ and $K$ GNN layers ($K$ iterations), GIN performs the $k$-th iterations of updating feature embedding $h^{(k)}$ for each node $v \in \mathcal{V}$:
\begin{equation}
     h^{(k)}_v=MLP^{(k)}((1+\epsilon^{(k)}) h^{(k-1)}_v + \sum_{u\in N_b(v)}h^{(k-1)}_u),
     \label{eq:GIN}
\end{equation}
where $h^{(k)}_v$ is the embedding of node $v$ at the $k$-th layer, $\epsilon^{(k)}$ is an arbitrary number that can be learned, and $N_b(v)$ is the neighbors of $v$ via edges in $\mathcal{E}$. 
Note that $h^{(0)}_v$ refers to its raw features for input. 
$MLP^{(k)}$ is a \textit{Multi-Layer Perceptron} (\textit{MLP}) for the $k$-th layer with a batch normalization \citep{GINBA_Ioffe2015BatchNA}.

Furthermore, we actually use \textit{heterogeneous GIN}, also called \textit{HGIN}, since there are two types of nodes, machine and operation nodes, and three relations, $O \to O$, $O \to M$ and $M \to O$ in the graph representation. 
Although we do not have cross machine relations $M \to M$ as described above, updating machine nodes requires to include the update from itself as in (\ref{eq:GIN}), that is, there is also one more relation $M \to M$. 
Thus, HGIN encodes graph information between all relations by using the four MLPs as follows,  
\begin{equation}
    h^{(k+1)}_v=\sum_{\mathcal{R}}MLP^{(k+1)}_{\mathcal{R}}((1+\epsilon^{(k+1)}_{\mathcal{R}})h^{(k)}_{v} + \sum_{u\in N_{\mathcal{R}}(v)}h^{(k)}_u)
\end{equation}
where $\mathcal{R}$ is one of the above four relations and $MLP^{(k)}_{\mathcal{R}}$ is the MLP for $\mathcal{R}$.
For example, for $S_0$ in Figure \ref{fig:partial_solution_steps} (a), the embedding of $M_1$ in the $(k+1)$-st iteration can be derived as follows. 
\kho{
\begin{equation}
    h^{(k+1)}_{M_1} 
     = MLP^{(k+1)}_{MM}((1+\epsilon^{(k+1)}_{MM})h^{(k)}_{M_1}) + MLP^{(k+1)}_{OM}( (h^{(k)}_{O_{1,1}}||\bar{T}_{1,1,1})+ (h^{(k)}_{O_{2,3}}||\bar{T}_{2,3,1}) + (h^{(k)}_{O_{3,1}}||\bar{T}_{3,1,1}))
\end{equation}
}
%
Similarly, the embedding of $O_{1,1}$ in the $(k+1)$-st iteration is: 
\kho{
\begin{equation}
     h^{(k+1)}_{O_{1,1}} 
     = MLP^{(k+1)}_{OO}((1+\epsilon^{(k+1)}_{OO})h^{(k)}_{O_{1,1}} + h^{(k)}_{O_{1,2}} + h^{(k)}_{O_{1,3}}) + MLP^{(k+1)}_{MO}((h^{(k)}_{M_{1}}||\bar{T}_{1,1,1}))
\end{equation}
}
%

In our approach, an action includes the two phases, graph embedding phase and action selection phase.
Let $h^{(k)}_{\mathcal{G}}$ denote the whole embedding of the graphs $\mathcal{G}$\kho{, including all $h^{(k)}_{O\in\mathcal{G}}$ and $h^{(k)}_{M\in\mathcal{G}}$}. 
In the graph embedding phase, we use an HGIN to encode node and graph embeddings as described above. 
An example with three HGIN layers is illustrated in Figure \ref{fig:networks_for_our_approach} (b). 

In the action selection phase, we select an action based on a policy, after node and graph embedding are encoded in the graph embedding phase. 
The policy is described as follows.
First, collect all \textit{ready} operations $O$ to be dispatched to machines $M$. 
Then, for all pairs ($M$, $O$), feed their node embeddings ($h^{(k)}_M$, $h^{(k)}_O$) into a MLP $Score(M,O)$ to calculate their scores as shown in Figure \ref{fig:networks_for_our_approach} (c).
The probability of selecting ($M$, $O$) is calculated based on a softmax function of all scores, which also serves as the model policy $\pi$ for the current state.

\subsection{Policy-Based RL Training}
\label{sec:training}

In this paper, we propose to use a policy-based RL training mechanism that follows REINFORCE \citep{REINFORCE_Sutton1998} to update our model by policy gradient with a normalized advantage makespan with respect to a baseline policy $\pi_b$ as follows. 
\begin{equation}
    \label{eq:NA_return}
    A_{\pi}(S,a) = \frac{T^{(mksp)}_{\pi_b}(S,a)-T^{(mksp)}_{\pi}(S,a)}{T^{(mksp)}_{\pi_b}(S,a)}
\end{equation}

In this paper, we choose a lightweight PDR, MWKR, as baseline $\pi_b$, which performed best for makespan among all PDRs reported from the previous work \citep{L2D_NIPS_ZhangSC0TX20}.
In fact, our experiment also shows that using MWKR is better than the other PDRs shown in the appendix. 
The model for policy $\pi$ is parametrized by $\theta$, which is updated by $\nabla_{\theta}log\pi_{\theta} A_{\pi_\theta}(S_t,a_t)$. 
Our algorithm based on REINFORCE is listed in the appendix.

\section{Experiments}
\label{sec:experiments}


\subsection{Experimental Settings and Evaluation Benchmarks}
\label{sec:experiment-settings}

In our experiments, the settings of our model are described as follows. 
All embedding and hidden vectors in our model have a dimension of 256. 
The model contains three HGIN layers for graph embedding, and an MLP for the score function, as shown in Figure \ref{fig:networks_for_our_approach} (b) and (c). 
All MLP networks including those in HGIN and for score contain two hidden layers.
The parameters of our model, such as MLP, generally follow the default settings in PyTorch \citep{pytorch_NEURIPS2019_9015} and PyTorch Geometric \citep{pytorchgeometric_Fey/Lenssen/2019}.
More settings are in the appendix. 

Each of our models is trained with one million episodes, each with one scheduling instance. 
Each instance is generated by following the procedure which is used to generate the TA dataset \citep{Ta_dataset_Taillard1993BenchmarksFB}.
Given ($N$, $M$), we use the procedure to generate an $n \times m$ JSP instance by conforming to the following distribution, $n \sim \mathcal{U}(3, N)$, $m \sim \mathcal{U}(3, n)$, and operation count $k_j=m$, where $\mathcal{U}(x,y)$ represents a distribution that uniformly samples an integer in a close interval $[x,y]$ at random.
The details of designation for machines and processing times refer to \citep{Ta_dataset_Taillard1993BenchmarksFB} and thus are omitted here. 
We choose (10,10) for all experiments, since (10,10) generally performs better than the other two as described in the appendix. 
Following the method described in Subsection \ref{sec:training}, the model is updated from the above randomly generated instances. 
For testing our models for JSP and FJSP, seven JSP open benchmarks and two FJSP open benchmarks are used, as listed in the appendix.

The performance for a given policy method $\pi$ on an instance is measured by the makespan gap $G$ defined as
\begin{equation}
    \label{eq:makespan_gap}
    G = \frac{T^{(mksp)}_{\pi}-T^{(mksp)}_{\pi*}}{T^{(mksp)}_{\pi*}}
\end{equation}
where $T^{(mksp)}_{\pi*}$ is the optimal makespan or the best-effort makespan, from a mathematical optimization tool, OR-Tools, serving as $\pi*$. 
By the best-effort makespan, we mean the makespan derived with a sufficiently large time limitation, namely half a day with OR-Tools.
For comparison in experiments, we use a server with Intel Xeon E5-2683 CPU and a single NVIDIA GeForce GTX 1080 Ti GPU.
Our method uses a CPU thread and a GPU to train and evaluate, while OR-Tools uses eight threads to find the solution.

\subsection{Experiments for JSP}
\label{sec:experiment_JSP}

For JSP, we first train a model based on residual scheduling, named RS. 
For ablation testing, we also train a model, named RS+op, by following the same training method but  without removing irrelevant operations. 
When using these models to solve testing instances, action selection is based on the greedy policy that simply chooses the action $(M,O)$ with the highest score deterministically, obtained from the score network as in Figure \ref{fig:networks_for_our_approach} (c).  

For comparison, we consider the three DRL construction heuristics, respectively developed in \citep{L2D_NIPS_ZhangSC0TX20} called L2D, \citep{GNNDRL_Park2021LearningTS} by Park et al., and \citep{ScheduleNet_abs-2106-03051}, called ScheduleNet.
We directly use the performance results of these methods for open benchmarks from their articles. 
For simplicity, they are named L2D, Park and SchN respectively in this paper. 
We also include some construction heuristics based PDR, such as MWKR, MOR, SPT and FIFO.
Besides, to derive the gaps to the optimum in all cases, OR-Tools serve as $\pi*$ as described in (\ref{eq:makespan_gap}). 

Now, let us analyze the performances of RS as follows. 
Table \ref{tab:makespanGapTA_noL2S} shows the average makespan gaps for each collection of JSP TA benchmarks with sizes, 15×15, 20×15, 20×20, 30×15, 30×20, 50×15, 50×20 and 100×20, where the best performances (the smallest gaps) are marked in bold. 
In general, RS performs the best, and generally outperforms the other methods for all collections by large margins, except for that it has slightly higher gaps than RS+op for the two collections, 15 $\times$ 15 and 20 $\times$ 20. 
In fact, RS+op also generally outperforms the rest of methods, except for that it is very close to SchN for two collections. 
For the other six open benchmarks, ABZ, FT, ORB, YN, SWV and LA, the performances are similar and thus presented in the appendix. 
It is concluded that RS generally performs better than other construction heuristics by large margins. 

\begin{table*}[tb!]
    \caption{Average makespan gaps for TA benchmarks.}
    \label{tab:makespanGapTA_noL2S}
    \centering
\begin{tabular}{c|c|c|c|c|c|c|c|c||c}
    \hline
    Size & 15$\times$15 & 20$\times$15 & 20$\times$20 & 30$\times$15 & 30$\times$20 & 50$\times$15 & 50$\times$20 & 100$\times$20 & $Avg.$ \\
    \hline
    \hline
    RS &  0.148 & \textbf{0.165} & 0.169 & \textbf{0.144} & \textbf{0.177} & \textbf{0.067} & \textbf{0.100} & \textbf{0.026} & \textbf{0.125} \\
    RS+op & \textbf{0.143} & 0.193 & \textbf{0.159} & 0.192 & 0.213 & 0.123 & 0.126 & 0.050 & 0.150 \\ 
    \hline
    MWKR & 0.191 & 0.233 & 0.218 & 0.239 & 0.251 & 0.168 & 0.179 & 0.083 & 0.195 \\
    MOR & 0.205 & 0.235 & 0.217 & 0.228 & 0.249 & 0.173 & 0.176 & 0.091 & 0.197 \\
    SPT  &  0.258 & 0.328 & 0.277 & 0.352 & 0.344 & 0.241 & 0.255 & 0.144 & 0.275 \\
    FIFO & 0.239 & 0.314 & 0.273 & 0.311 & 0.311 & 0.206 & 0.239 & 0.135 & 0.254 \\
    \hline
    \hline
    L2D & 0.259 & 0.300 & 0.316 & 0.329 & 0.336 & 0.223 & 0.265 & 0.136 & 0.270\\
    Park & 0.201 & 0.249 & 0.292 & 0.246 & 0.319 & 0.159 & 0.212 & 0.092 & 0.221 \\
    SchN & 0.152 & 0.194 & 0.172 & 0.190 & 0.237 & 0.138 & 0.135 & 0.066 & 0.161 \\
    \hline
\end{tabular}
\end{table*}


\subsection{Experiments for FJSP}
\label{sec:experiment_FJSP}

\begin{table*}[h]
    \caption{Average makespan gaps for FJSP open benchmarks}
    \label{tab:FJSP_public_dataset_greedy}
    \centering
\begin{tabular}{c|c|c|c|c}
    \hline
    Method & MK & LA(rdata) & LA(edata) & LA(vdata)  \\
    \hline
    \hline
    RS & \textbf{0.232} & \textbf{0.099} & \textbf{0.146} & 0.031   \\
    RS+op & 0.254 & 0.113 & 0.168 & \textbf{0.029} \\ 
    \hline
    DRL-G & 0.254 & 0.111 & 0.150  & 0.040 \\
    \hline
    MWKR &  0.282 &  0.125 & 0.149 & 0.051 \\
    MOR & 0.296  & 0.147 & 0.179 & 0.061 \\
    SPT & 0.457  & 0.277 & 0.262 & 0.182 \\
    FIFO & 0.307 & 0.166 & 0.220 & 0.075 \\
    \hline
\end{tabular}
\end{table*}

For FJSP, we also train a model based on residual scheduling, named RS, and an ablation version, named RS+op, without removing irrelevant operations. 
We compares ours with one DRL construction heuristics developed by \citep{FJSP_L2D_SongCLC23}, called DRL-G, and four PDR-based heuristics, MOR, MWKR, SPT and FIFO.
We directly use the performance results of these methods for open datasets according to the reports from \citep{FJSP_L2D_SongCLC23}. 


Table \ref{tab:FJSP_public_dataset_greedy} shows the average makespan gaps in the four open benchmarks, MK, LA(rdata), LA(edata) and LA(vdata). 
From the table, RS generally outperforms all the other methods for all benchmarks by large margins, except for that RS+op is slightly better for the benchmark LA(vdata). 

\section{Discussions}
\label{sec:large-instances}
\label{sec:discussion}

In this paper, we propose a new approach, called residual scheduling, to solving JSP an FJSP problems, and the experiments show that our approach reaches SOTA among DRL-based construction heuristics on the above open JSP and FJSP benchmarks. 
We further discusses three issues: large instances, computation times and further improvement. 

\begin{figure}[h]
    \centering

    \includegraphics[width=1.0\textwidth]{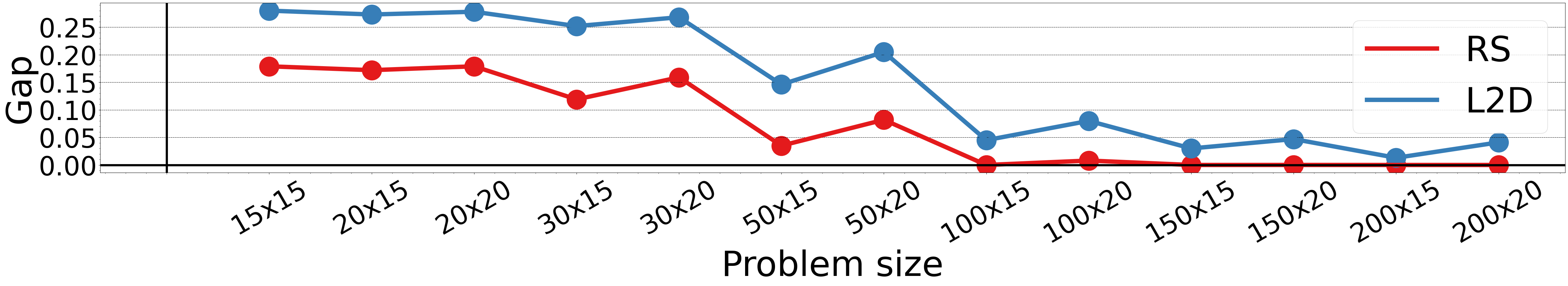}

    \caption{Average makespan gaps of JSP instances with different problem sizes. }
    \label{fig:large_JSP_case}
\end{figure}

First, from the above experiments particularly for TA benchmark for JSP, we observe that the average gaps gets smaller as the number of jobs increases, even if we use the same model trained with $(N,M)=(10,10)$.
In order to investigate size-agnostics, we further generate 13 collections of JSP instances of sizes for testing, from 15 $\times$ 15 to 200 $\times$ 20, and generate 10 instances for each collection by using the procedure above. 
Figure \ref{fig:large_JSP_case} shows the average gaps for these collections for RS and L2D, and these collections are listed in the order of sizes in the x-axis. 
Note that we only show the results of L2D in addition to our RS, since L2D is the only open-source among the above DRL heuristics. 
\kho{In Figure 4, it is observed that RS is much better than L2D.}
Interestingly, using RS, the average gaps are nearly zero for the collections with sizes larger than 100 $\times$ 15, namely, 100 $\times$ 15, 100 $\times$ 20, 150 $\times$ 15, 200 $\times$ 15 and 200 $\times$ 20. 
Among the 50 JSP instances in the five collections, 49 reaches \kho{zero makespan gaps}. 
A strong implication is that our RS approach can be scaled up for job sizes and even reach the optimal for sufficient large job count. 

Second, the computation times for RS are relatively small and has low variance like most of other construction heuristics.
Here, we just use the collection of TA 100x20 for illustration. 
It takes about 30 seconds on average for both RS and RS+op, about 28 for L2D and about 444 for SchN. 
In contrast, it takes about 4000 seconds with high variance for OR-Tools. 
The times for other collections are listed in more detail in the appendix. 

\begin{table*}[h]
    \caption{Average makespan gaps for FJSP open benchmark.}
\label{tab:FJSP_public_dataset_sample}
    \centering
\begin{tabular}{c|c|c|c|c}
    \hline
    Method & MK & LA(rdata) & LA(edata) & LA(vdata)  \\
    \hline
    \hline
    RS & 0.232 & 0.099 & 0.146 & 0.031 \\
    RS+100 & \textbf{0.154} & \textbf{0.047} & \textbf{0.079} & \textbf{0.007}\\ 
    \hline
    DRL-G & 0.254 & 0.111 & 0.150 & 0.040\\
    DRL+100 & 0.190 & 0.058 & 0.082 & 0.014\\
    \hline
\end{tabular}
\end{table*}

Third, as proposed by \citet{FJSP_L2D_SongCLC23}, construction heuristics can further improve the gap by constructing multiple solutions based on the softmax policy, in addition to the greedy policy. 
They had a version constructing 100 solutions for FJSP,  called DRL+100 in this paper. 
In this paper, we also implement a RS version for FJSP based on the softmax policy, as described in Subsection \ref{sec:gnn}, and then use the version, called RS+100, to constructing 100 solutions.  
In Table \ref{tab:FJSP_public_dataset_sample}, the experimental results show that RS+100 performs the best, much better than RS, DRL-G and DRL+100. 
An important property for such an improvement is that constructing multiple solutions can be done in parallel. 
That is, for construction heuristics, the solution quality can be improved by adding more computation powers.

\bibliographystyle{ACM-Reference-Format}
\bibliography{NIPS_main}

\newpage
\appendix
\section*{Appendix}

In this appendix, we include the following items, described in the subsequent subsections respectively. 
\begin{itemize}
    \item Code, datsets and model weights. 
    \item Formulation of Markov decision process for solution construction. 
    \item Illustration of graph representation for partial solutions in Figure \ref{fig:partial_solution_steps}. 
    \item REINFORCE algorithm in our approach. 
    \item Experimental settings.
    \item Additional Experiments. 
\end{itemize}

\section{Code, Datasets and Model Weights}
\label{app:code_and_dataset}
Our code, datasets and pre-trained weights are in the supplementary folder {\tt ResidualScheduling}.
The code includes the following: 
First, setup the virtual environment, 
Then, install required packages in the environment.
Finally, execute the code with the commands in the {\tt README} file.
All hyperparameters are also listed in {\tt README} file.

We collect the following dataset for evaluation.
For JSP dataset, we collect 13 collections generated for the experiments shown in Figure \ref{fig:large_JSP_case}, and other seven JSP benchmarks, TA, ABZ, FT, ORB, YN, SWV and LA.
For FJSP dataset, we collect two FJSP benchmarks, MK and LA, and the instances generated from the work \citep{FJSP_L2D_SongCLC23}.

There are four weights according to the final models for RS and RS+op with JSP and FJSP cases, denoted as {\tt RS\_JSP}, {\tt RS+op\_JSP}, {\tt RS\_FJSP}, and {\tt RS+op\_FJSP}.

\section{Formulation of Markov Decision Process for Solution Construction}
\label{sec:MDPformulation}

In Subsection \ref{sec:constuction}, an approach based on solution construction can be also viewed as the so-called \textit{Markov decision process} (\textit{MDP}).
An MDP is a stochastic decision-making process widely used in reinforcement learning, generally defined 
by a tuple $(\mathcal{S}, \mathcal{A}, \mathcal{R}, \mathcal{P})$, where $\mathcal{S}$ is the finite set of the states, $\mathcal{A}$ is the set of the available actions, $\mathcal{R}$ is the reward function and $\mathcal{P}$ is the transition probability function.
The objective is to find a policy that maximizes the agent's cumulative rewards\footnote{Discount factor $\gamma$ is not included in the tuple, since it is always one and thus not used in this paper.}.

Following the above MDP definition, we formulate the process of solution construction for JSP and FJSP problems as follows. 
\begin{itemize}
    \item $\mathcal{S}$ specifies states each representing an instance associated with the information of partial solutions $S$ like those in Figure \ref{fig:partial_solution_steps}.
    \item $\mathcal{A}$ specifies actions each of which selects machine-operation pairs ($M,O$) to dispatch the operation $O$ on the machine $M$ on given states.
    \item $\mathcal{R}$ specifies a reward to indicate a negative of the additional processing time for the dispatched action. 
    \item $\mathcal{P}$ specifies how states are transited from one partial solution $S$ to the next $S'$ after an action, as described in Subsection \ref{sec:constuction} accordingly.
\end{itemize}
An episode starts from the initial partial solution and repeats transitions till the end when the partial solution is a solution.
The cumulative reward is the negative of makespan $T^{(mksp)}$, i.e., the total complete time.
The objective is to maximize the cumulative reward, i.e., minimize the makespan. 
In the past, many methods for JSP and FJSP based on reinforcement learning such as \citep{KoreaRLJSP_Park2020ARL,BeijingNTHU_Luo2020DynamicSF,L2D_NIPS_ZhangSC0TX20,GNNDRL_Park2021LearningTS,ScheduleNet_abs-2106-03051} followed this MDP formulation.

\section{Illustration of Graph Representation for Partial Solutions }
\label{sec:graph-illustration}
\begin{figure*}[ht]
    \centering
    \subfigure[$S_0$]{
    \includegraphics[width=0.3\textwidth]{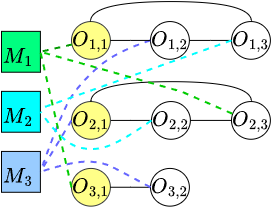}
    \label{fig:JSP_solution_graph_0}
    }
    \subfigure[$S_1$]{
    \includegraphics[width=0.3\textwidth]{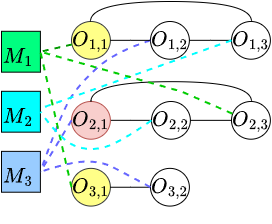}
    \label{fig:JSP_solution_graph_1}
    }
    \subfigure[$S_2$]{
    \includegraphics[width=0.3\textwidth]{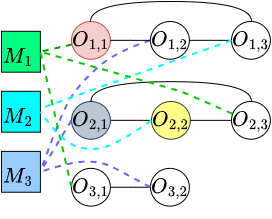}
    \label{fig:JSP_solution_graph_2}
    }
    \subfigure[$S_3$]{
    \includegraphics[width=0.3\textwidth]{Figures/partial_solution/partial_step3_v1.png}
    \label{fig:JSP_solution_graph_3_appendix}
    }
    \subfigure[$S_4$]{
    \includegraphics[width=0.3\textwidth]{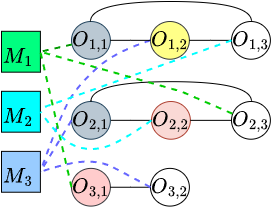}
    \label{fig:JSP_solution_graph_4}
    }
    \subfigure[$S_5$]{
    \includegraphics[width=0.3\textwidth]{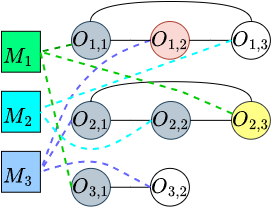}
    \label{fig:JSP_solution_graph_5}
    }
    \subfigure[$S_6$]{
    \includegraphics[width=0.3\textwidth]{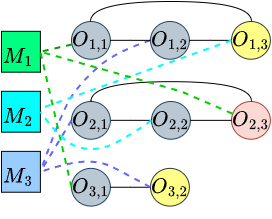}
    \label{fig:JSP_solution_graph_6}
    }
    \subfigure[$S_7$]{
    \includegraphics[width=0.3\textwidth]{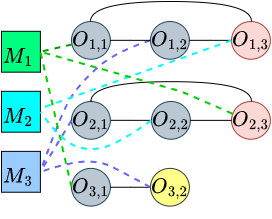}
    \label{fig:JSP_solution_graph_7}
    }
    \subfigure[$S_8$]{
    \includegraphics[width=0.3\textwidth]{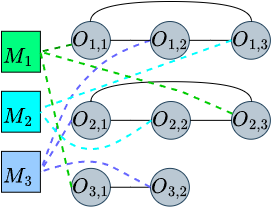}
    \label{fig:JSP_solution_graph_8}
    }
    \caption{(a)-(i) are corresponding to partial solutions in Figure \ref{fig:partial_solution_steps} (a)-(i), respectively.}
    \label{fig:partial_solution_graph_steps}
\end{figure*}

In Figure \ref{fig:networks_for_our_approach} (a), we illustrate graph representation of $S_3$. 
In this section, Figure \ref{fig:partial_solution_graph_steps} illustrates the step-by-step graph representations respectively corresponding to all the partial solutions in Figure \ref{fig:partial_solution_steps} in Subsection \ref{sec:residual-graph-rep}.
Again, the gray nodes are \emph{completed}, the red \emph{ongoing}, the yellow \emph{ready} and the white \emph{unready}.

\section{REINFORCE Algorithm in Our Approach}


Our residual scheduling approach basically follows REINFORCE \citep{REINFORCE_Sutton1998} to update our model by policy gradient with a normalized advantage makespan based on a baseline policy and an entropy bonus to ensure sufficient exploration like PPO. 
Algorithm \ref{algo:REINFORCE} (next page) shows our REINFORCE training process with the normalized advantage makespan described in Section \ref{sec:training}. 
The model policy $\pi_{\theta}$ is parameterized by $\theta$ initialized at random, and a baseline policy $\pi_{b}$ is given, which is a lightweight PDR, MWKR, in this paper. 
The learning rate $\alpha \in (0, 1)$ is initialized to $10^{-4}$ and decayed by a factor of 0.99 every $P=1000$ episodes in this paper.

\begin{algorithm}[h!]
\caption{REINFORCE with a normalized advantage for JSP and FJSP}
\label{algo:REINFORCE}
\begin{algorithmic}[1]
 \STATE {\bfseries Input:} Initial policy 
    $\pi_{\theta}$, learning rate $\alpha\in(0, 1)$, entropy bonus coefficient $c$, a baseline policy $\pi_b$;
 \FOR{Episode $e=0, 1, \cdots$}
    \STATE Use policy $\pi_{\theta}$ to rollout an episode: $\{S_1, a_1, S_2, a_2, \cdots, S_T \}$
    \FOR{$t=0, 1, 2, \cdots T$}
      \STATE Use the baseline policy $\pi_b$ to rollout and obtain $T^{(mksp)}_{\pi_b}(S_t,a_t)$
      \STATE Calculate $T^{(mksp)}_{\pi_\theta}(S_t,a_t)$ 
      \STATE Calculate normalized advantage $A_{\pi_\theta}(S_t,a_t)$ following Equation (\ref{eq:NA_return})
      \STATE Update policy
      $\theta=\theta+\alpha\nabla_{\theta}(log\pi_{\theta} A_{\pi_\theta}(S_t,a_t)+cH_{\pi_\theta}(S_t) ) $
    \ENDFOR
    \STATE Update learning rate every $P=1000$ episodes 
 \ENDFOR
\end{algorithmic}
\end{algorithm}

\section{Experimental Settings}
\label{app:experiment_details}

In this section, we first describe settings of our model in Appendix \ref{app:model_settings} and then list the details of JSP and FJSP open benchmarks used in our experiments in Appendix \ref{sec:open-benchmarks}.

\subsection{Model Settings}
\label{app:model_settings}

In this subsection, we first list the settings of our model in Table \ref{tab:Model_architecture}.
Particularly, the entropy bonus coefficient $c$ is used for the weight of entropy in line 8 of Algorithm \ref{algo:REINFORCE}, and the initial learning rate $\alpha$ is used in lines 1 and 8 of Algorithm \ref{algo:REINFORCE}. 

\begin{table*}[htb!]
    \caption{Model hyperparameters.}
    \label{tab:Model_architecture}
    \centering
    \begin{tabular}{|p{0.2\textwidth}|c|p{0.6\textwidth}|}
    \hline
        Item & Value & Description\\
        \hline
        \hline
        HGIN layers &  3 & Layer number of GNN.\\
        \hline
         FC layers of MLP & 2 & Layer number of MLP.\\
        \hline
        Hidden dimension & 256 & Hidden vector dimension for GNN and policy network.\\
        \hline
        Episodes & 1,000,000 & Training episodes\\
        \hline
        Entropy bonus coefficient ($c$) & $10^{-2}$ & The weight of entropy bonus in the loss function.\\
        \hline
        Initial learning rate ($\alpha$) & $10^{-4}$ & Initialized with $10^{-4}$ and then decayed with 0.99 rate for every 1000 episodes\\
        \hline
        Optimizer & Adam & Adam with $\beta_1 = 0.9$ and $\beta_2 = 0.999$\\
        
    \hline
    \end{tabular}
\end{table*}

In practice, one model is trained for each of $(N,M)$, (6,6), (10,10) and (15,15), and each of baseline policies, MWKR, MOR, SPT and FIFO. 
Each of models is trained with 1000 iterations, for each of which model parameters are updated with 1000 episodes (one million of episodes in total). \textcolor{blue}{It takes about one day to train with 200,000 episodes.}
For each episode, one scheduling instance is generated for training by the procedure subject to $(N,M)$ as in Subsection \ref{sec:experiment-settings}.

For JSP, we also generate a validation dataset of the three collections, 20 $\times$ 20, 50 $\times$ 20 and 100 $\times$ 20, each with 10 instances. 
Among the 1000 sets of model parameters, the one performing best for the validation dataset is chosen for testing in the rest of experiments.
Table \ref{tab:ablation_baseline_policy_validation} shows the average makespans of the four models trained with different baselines respectively and with (10,10) on the validation dataset. 
The one with MWKR clearly outperforms other baselines on the validation dataset. 
Thus, MWKR is chosen as the baseline in this paper.
Similarly, Table \ref{tab:ablation_training_size_JSPvalidation} shows the average makespans of three models trained with different training sizes (6,6), (10,10) and (15,15) respectively and with MWKR on the validation dataset. 
The one with (10,10) performs best, except for that the one with (15,15) performs slightly better for the collection of 100 $\times$ 20.
Thus, this paper still chooses (10,10) since it performs well in general and has less training costs when compared to (15,15). 

\begin{table*}[htb!]
    \caption{Average makespans of models with different baseline policies $\pi_b$ on JSP validation dataset.}
    \label{tab:ablation_baseline_policy_validation}
    \centering
\begin{tabular}{c|c|c|c}
    \hline
    $\pi_b$ & 20$\times$20 & 50$\times$20 & 100$\times$20 \\
    \hline
    \hline
    MWKR & \textbf{1803.1} & \textbf{3147.3} & \textbf{5676.0} \\
    MOR  & 1831.7 & 3229.8 & 5728.3 \\
    SPT  & 1813.8 & 3201.7 & 5718.7 \\
    FIFO & 1826.4 & 3177.6 & 5692.9 \\
    \hline
\end{tabular}\
\end{table*}

\begin{table*}[htb!]
    \caption{Average makespans of models with different $(N,M)$ on JSP validation dataset.}
    \label{tab:ablation_training_size_JSPvalidation}
    \centering
\begin{tabular}{c|c|c|c}
    \hline
    Method & 20$\times$20  & 50$\times$20 & 100$\times$20  \\
    \hline
    \hline
    RS (6,6)  & 1822.8 & 3242.7 & 5784.4  \\
    RS (10,10)  & \textbf{1803.1} & \textbf{3147.3} & 5676.0  \\
    RS (15,15)  & 1820.7 & 3170.6 & \textbf{5670.4}  \\
    \hline
\end{tabular}\
\end{table*}
For FJSP, we simply use the validation set used by \citep{FJSP_L2D_SongCLC23}. 
Table \ref{tab:ablation_FJSP_public_dataset_datasize} shows the average makespans of three models trained with different training sizes (6,6), (10,10) and (15,15) respectively and with MWKR on the validation dataset. 
The one with (10,10) outperforms others for all the collections in \cite{FJSP_L2D_SongCLC23}. 
Hence, (10,10) is also chosen in FJSP experiments in this paper. 

\begin{table*}[h!]
    \caption{Average makespans of models with different $(N,M)$ on FJSP validation dataset.}
\label{tab:ablation_FJSP_public_dataset_datasize}
    \centering
\begin{tabular}{c|c|c|c|c}
    \hline
    Method & 10 $\times$ 5 & 15 $\times$ 10 & 20 $\times$ 5 &  20 $\times$ 10\\
    \hline
    \hline
    RS (6,6)   &  106.25 & 164.98 & 209.77 & 210.57 \\
    RS (10,10) &  \textbf{106.00} & \textbf{163.88} & \textbf{209.07} & \textbf{210.18} \\
    RS (15,15) &  106.15 & 163.95 & 209.57 & 210.47 \\
    \hline
\end{tabular}
\end{table*}

\subsection{Open Benchmarks}
\label{sec:open-benchmarks}

Table \ref{tab:benchmark_info} lists the information about JSP and FJSP open benchmarks used in this paper.
For JSP, TA benchmark has most instances, including 8 categories, 15 $\times$ 15, 20 $\times$ 15, 20 $\times$ 20, 30 $\times$ 15, 30 $\times$ 20, 50 $\times$ 15,  50 $\times$ 20 and 100 $\times$ 20.
For each category, there are 10 instances generated by their procedure.
For other JSP benchmarks, they cover different problem sizes and data distributions as in the table.

For FJSP, two popular open benchmarks are MK, also known as Brandimarte instances, and LA, also known as Hurink instances (to distinguish LA benchmark for JSP).
Hurink instances are categories into three types, edata, rdata and vdata, according to different distribution for assignable machines, denoted by LA(edata), LA(rdata) and LA(vdata) respectively, as described in \citep{FJSP_LA_dataset_hurink1994tabu}.

\begin{table}[h!]
    \caption{Open benchmarks.}
    \label{tab:benchmark_info}
    \centering
\begin{tabular}{c|l|ccc}
    \hline
    Type & Benchmark & Instances & Min size & Max size\\
    \hline
    \hline
    \multirow{7}{*}{JSP} &
    TA \citep{Ta_dataset_Taillard1993BenchmarksFB} & 80 & 15$\times$15 & 100$\times$20 \\
    \cline{2-5}
    & ABZ \citep{ABZ_dataset_Adams1988TheSB}
    & 5 & 10$\times$10 & 20$\times$15 \\
    \cline{2-5}
    & FT \citep{FT_dataset_muth1963industrial} & 3 & 6$\times$6 & 20$\times$5 \\
    \cline{2-5}
    & ORB \citep{ORB_dataset_Applegate1991ACS} & 10 & 10$\times$10 & 10$\times$10 \\
    \cline{2-5}
    & YN \citep{YN_dataset_Yamada1992AGA} & 4 & 20$\times$20 & 20$\times$20 \\
    \cline{2-5}
    & SWV \citep{SWV_dataset_Storer1992NewSS} & 20 & 20$\times$10 & 50$\times$10 \\
    \cline{2-5}
    & LA \citep{LA_dataset} & 40 & 10$\times$5 & 30$\times$10 \\
    \hline
    \multirow{2}{*}{FJSP} &
    MK \citep{MK_dataset_Brandimarte93} & 10 & 10$\times$6 & 20$\times$15 \\
    \cline{2-5}
    & LA \citep{FJSP_LA_dataset_hurink1994tabu}
    & 120 & 10$\times$5 & 30$\times$10 \\
    \hline
\end{tabular}
\end{table}

\section{Additional Experiments}
\label{app:additional_experiments_results}

This section shows additional experiments conducted to support residual scheduling as follows.
\begin{itemize}
    \item Appendix \ref{app_sec:comaprsion_otherJSP} shows the comparisons between RS and other construction heuristics for other six JSP benchmarks. 
    \item Appendix \ref{app_sec:comaprsion_randomFJSP} shows the comparisons between RS and other methods for FJSP instances generated by \citet{FJSP_L2D_SongCLC23}. 
    \item Appendix \ref{app_sec:computation_time} shows the computation times of construction heuristics for comparison. 
    \item Appendix \ref{app:experiments_results} shows the makespans of all instances in more detail. 
\end{itemize}

\subsection{Other JSP Benchmarks}
\label{app_sec:comaprsion_otherJSP}

Table \ref{tab:makespanGapOtherDataset_noL2S} shows the average makespan gaps for other six JSP open benchmarks, where the best performances (the smallest gaps) are marked in bold. 
Similar to the results in TA benchmark, in general, RS still performs the best, and generally outperforms the other methods except for that RS has slightly higher gaps than RS+op for the ORB benchmark. 
In fact, RS+op also generally outperforms the rest of methods, except for that RS+op is slightly higeher than but very close to SchN for ABZ and YN benchmarks. 
It is also concluded that RS generally performs better than other construction heuristics.

\begin{table*}[h!]
    \caption{Average makespan gaps for other benchmarks. }
    \label{tab:makespanGapOtherDataset_noL2S}
    \centering
\begin{tabular}{c|c|c|c|c|c|c}
    \hline
    Dataset & ABZ & FT & ORB & YN & SWV & LA \\
    \hline
    \hline
    RS &  \textbf{0.116} & \textbf{0.089} & 0.165 & \textbf{0.166} & \textbf{0.122} & \textbf{0.075} \\
    RS+op & 0.156 & 0.162 & \textbf{0.159} & 0.186 & 0.202 &  0.097 \\
    \hline
    MWKR & 0.166 & 0.196 & 0.251 & 0.197 & 0.284 & 0.126\\    
    MOR & 0.180 & 0.232 & 0.290 & 0.228 & 0.358 & 0.138 \\
    SPT & 0.251 & 0.280 & 0.262 & 0.306 & 0.230 & 0.199 \\
    FIFO & 0.247 & 0.288 & 0.297 & 0.258 & 0.373 & 0.188 \\
    \hline
    Park & 0.214 & 0.222 & 0.218 & 0.247 & 0.228 & 0.141 \\
    SchN & 0.147 & 0.184 & 0.199 & 0.184 & 0.288 & 0.099 \\
    \hline
\end{tabular}
\end{table*}

\subsection{FJSP Dataset from Song et al.}
\label{app_sec:comaprsion_randomFJSP}

This subsection also compares our model with others on the FJSP instances generated by \citet{FJSP_L2D_SongCLC23}, as listed in Tables \ref{tab:FJSP_random_dataset_makespn_greedy} and \ref{tab:FJSP_random_dataset_makespn_sample}, where there are 100 FJSP instances for each problem size.
The two tables also show the makespan results for RS, RS+100, DRL-G, DRL+100, MOR, MWKR, SPT, FIFO and OR-Tools, since they only provided the average makespans, not makespan for each instance. 
Note that the average makespan gap is not able to be calculated with Equation (\ref{eq:makespan_gap}) without makespans from individual instances.
Besides, for OR-Tools, their results are taken 1800 seconds to find the best-effort solutions.
From Table \ref{tab:FJSP_random_dataset_makespn_greedy}, RS outperforms all the other construction heuristics, including DRL-G, for all problem sizes. 
From Table \ref{tab:FJSP_random_dataset_makespn_sample}, RS+100 clearly outperforms DRL+100, DRL-G and RS by significant margins, like Table \ref{tab:FJSP_public_dataset_sample} in Subsection \ref{sec:discussion}. 

\begin{table*}[htb!]
    \caption{Average makespans for the FJSP dataset by Song et al.}
\label{tab:FJSP_random_dataset_makespn_greedy}
    \centering
\begin{tabular}{c|c|c|c|c|c|c}
    \hline
    Method & 10 $\times$ 5 & 15 $\times$ 10 & 20 $\times$ 05 & 20 $\times$ 10 & 30 $\times$ 10 & 40 $\times$ 10 \\
    \hline
    \hline
    OR-Tools & 96.59 & 145.42 & 188.45 & 197.24 & 294.1 & 397.36\\
    \hline
    RS & \textbf{109.06} & \textbf{165.45} & \textbf{207.27} & \textbf{213.12} & \textbf{311.16} & \textbf{413.46} \\
    \hline
    DRL-G  &   111.67 & 166.92 & 211.22	& 215.78 & 313.04 & 416.18 \\
    \hline
    MOR    &  116.69 & 173.40 & 217.17 & 221.86 & 320.18 & 425.19  \\
    MWKR   &  115.29 & 169.18 & 216.98 & 220.85 & 319.89 & 425.70  \\
    SPT    &  129.06 & 198.20 & 229.89 & 254.59 & 347.40 & 443.30  \\
    FIFO   &  119.62 & 185.08 & 216.13 & 234.21 & 328.50 & 427.22  \\
    \hline
\end{tabular}
\end{table*}

\begin{table*}[htb!]
    \caption{Average makespans for the FJSP dataset by Song et al.}
    \label{tab:FJSP_random_dataset_makespn_sample}
    \centering
\begin{tabular}{c|c|c|c|c|c|c}
    \hline
    Method & 10 $\times$ 5 & 15 $\times$ 10 & 20 $\times$ 05 & 20 $\times$ 10 & 30 $\times$ 10 & 40 $\times$ 10 \\
    \hline
    \hline
    OR-Tools & 96.59 & 145.42 & 188.45 & 197.24 & 294.1 & 397.36\\
    \hline
    RS & 109.06 & 165.45 & 207.27 & 213.12 & 311.16 & 413.46 \\
    RS+100 & \textbf{103.31} & \textbf{157.30} & \textbf{201.73} & \textbf{207.86} & \textbf{305.27} & \textbf{407.28} \\
    \hline
    DRL-G  &   111.67 & 166.92 & 211.22	& 215.78 & 313.04 & 416.18 \\
    DRL+100  &   105.61 & 160.36 & 207.50	& 214.87 & 312.20 & 415.14 \\
    \hline
\end{tabular}
\end{table*}

\subsection{Computation Time}
\label{app_sec:computation_time}


Table \ref{tab:computationTime_TA_noL2S} and Table \ref{tab:computationTime_FJSP_public_dataset_greedy} show the computation time taken by methods for TA JSP benchmark and FJSP benchmarks.
The four heuristics, MWKR, MOR, SPT and FIFO, take almost the same time to obtain the scheduling solutions, so they are merged into the PDRs category.
For TA benchmark, the computation times for L2D and SchN were reported by their works \citep{L2D_NIPS_ZhangSC0TX20} and \citep{ScheduleNet_abs-2106-03051}, respectively, which also reported to use a machine with AMD Ryzen 3600 CPU and a single Nvidia GeForce 2070S GPU.
To our knowledge, there were no time records available for the work \citep{GNNDRL_Park2021LearningTS}, marked as "-".
We observe that the time for RS is much less than SchN, and comparable to other DRL-based construction heuristics.

\begin{table*}[htb!]
    \caption{Average computation times for TA JSP instances.}
    \label{tab:computationTime_TA_noL2S}
    \centering
\begin{tabular}{c|c|c|c|c|c|c|c|c||c}
    \hline
    Size & 15$\times$15 & 20$\times$15 & 20$\times$20 & 30$\times$15 & 30$\times$20 & 50$\times$15 & 50$\times$20 & 100$\times$20 & $Avg.$ \\
    \hline
    \hline
    RS & 0.47s & 0.83s & 0.91s & 1.93s & 2.21s & 5.30s & 6.96s & 27.32s & 5.74s\\ 
    \hline
    PDRs & 0.05s & 0.08s & 0.12s & 0.16s & 0.24s & 0.41s & 0.60s & 2.20s & 0.48s  \\ 
    \hline
    L2D
    & 0.40s & 0.60s & 1.10s & 1.30s & 1.50s & 2.20s & 3.6s & 28.20s & 4.86s \\ 
    \hline
    Park
    & - & - & - & - & - & - & - & - & - \\ 
    \hline
    SchN 
    & 3.50s & 6.60s & 11.00s & 17.10s & 28.30s & 52.50s & 96.00s & 444.0s & 82.38s \\ 
    \hline
\end{tabular}
\end{table*}

\begin{table*}[htb!]
    \caption{Average computation times for FJSP open benchmarks.}
    \label{tab:computationTime_FJSP_public_dataset_greedy}
    \centering
\begin{tabular}{c|c|c|c|c}
    \hline
    Method & MK & LA(rdata) & LA(edata) & LA(vdata)  \\
    \hline
    \hline
    RS & 0.85s & 0.63s & 0.80s & 0.98s \\ 
    \hline 
    PDRs & 0.41s & 0.47s & 0.47s & 0.46s \\
    \hline
    DRL-G & 0.90s & 0.97s & 1.04s & 0.96s\\
    \hline
\end{tabular}
\end{table*}

\subsection{Experiment Results in More Detail}
\label{app:experiments_results}

This section presents experiment results in more detail by listing makespans for all instances. 
First, Table \ref{tab:app_details_Large_dataset_makespan} shows the average JSP makespan gaps in Figure \ref{fig:large_JSP_case}, and Tables \ref{tab:app_Large_dataset_makespan1} and \ref{tab:app_Large_dataset_makespan2} show in more detail the makespans of all individual JSP instances obtained by RS, L2D and OR-tools.
For OR-tools, the time limitation is set to half a day, i.e., it is set to the best-effort if OR-tools exceed the time limitation, where the cases are marked with \dag. 
For all 60 JSP instances larger than and equal to 100 $\times$ 15, RS reaches all optimum (marked in bold) except for four 100 $\times$ 20 instances and one 150 $\times$ 20 instance in Table \ref{tab:app_Large_dataset_makespan2}.

Second, Tables \ref{tab:app_FJSP_MK_datasetDetails} and \ref{tab:app_FJSP_LA_datasetDetails} show the makespans of FJSP MK instances and LA instances, respectively.
OPT indicates the best-known results from \citep{Best_known_Solution_behnke2012test}.
From the tables, RS+100 obtains makespans the same as (marked in bold) or close to those in OPT for many instances.

Third, Table \ref{tab:app_TAdataset_makespan} and Table \ref{tab:app_TAdataset_makespan2}   show the makespans of all TA instances for Table \ref{tab:makespanGapTA_noL2S}.
Park and SchN represent the work \cite{GNNDRL_Park2021LearningTS} and ScheduleNet; and OPT is the best-effort results for OR-tools in 6000 seconds.
To indicate the best performance of the construction heuristics, we mark in bold the makespan that is closest to that of OPT.
We observe that RS performs best for most instances from Ta30 to Ta80, and outperforms others for nearly all the instances with sizes larger than and equal to  30 $\times$ 15.

\begin{table*}[htb!]
    \caption{Average makespan gaps in Figure \ref{fig:large_JSP_case}. }
    \label{tab:app_details_Large_dataset_makespan}
    \centering
    \begin{tabular}{c|c|c|c}
    \hline
    Method & OR-Tools & RS & L2D \\
    \hline
    \hline
    15 $\times$ 15 & 0 & 0.169 & 0.280 \\
    20 $\times$ 15 & 0\dag & 0.172 & 0.273 \\
    20 $\times$ 20 & 0\dag & 0.179 & 0.278 \\
    30 $\times$ 15 & 0\dag & 0.119 & 0.252 \\
    30 $\times$ 20 & 0\dag & 0.159 & 0.268 \\
    50 $\times$ 15 & 0 & 0.035 & 0.146 \\
    50 $\times$ 20 & 0\dag & 0.082 & 0.205 \\
    100 $\times$ 15 & 0 & 0.000 & 0.045 \\
    100 $\times$ 20 & 0 & 0.008 & 0.080 \\
    150 $\times$ 15 & 0 & 0.000 & 0.030 \\
    150 $\times$ 20 & 0 & 0.00005 & 0.047 \\
    200 $\times$ 15 & 0 & 0.000 & 0.013 \\
    200 $\times$ 20 & 0 & 0.000 & 0.041 \\
    \hline
    \end{tabular}
    \\
    \begin{minipage}{10cm}
    \small \dag means the best-effort solution from OR-Tools with a half day computation.
    \end{minipage}
\end{table*}

\begin{table*}[htb!]
    \centering
    \caption{Part 1 of makespans of all JSP instances of Table \ref{tab:app_details_Large_dataset_makespan}.}
    \label{tab:app_Large_dataset_makespan1}

\begin{tabular}{c|c|c|c|c||c|c|c|c}
    \hline
    Instance  & $n \times m$ & L2D & RS & OR-Tools & $n \times m$ & L2D & RS & OR-Tools\\
    \hline
    \hline
    01 & $15\times15$ & 1528  & 1386  &  1181 & - & - & - & -\\    
    02 & $15\times15$ & 1422  & 1360  &  1172 & - & - & - & - \\
    03 & $15\times15$ & 1500  & 1494  &  1243 & - & - & - & -\\
    04 & $15\times15$ & 1651  & 1444  &  1230 & - & - & - & -\\
    05 & $15\times15$ & 1663  & 1502  &  1302 & - & - & - & -\\
    06 & $15\times15$ & 1479  & 1399  &  1237 & - & - & - & -\\
    07 & $15\times15$ & 1590  & 1363  &  1139 & - & - & - & -\\
    08 & $15\times15$ & 1432  & 1343  &  1148 & - & - & - & -\\
    09 & $15\times15$ & 1591  & 1371  &  1170 & - & - & - & -\\
    10 & $15\times15$ & 1492  & 1352  &  1168 & - & - & - & -\\
    \hline    
    01 & $20\times15$ & 1754  & 1438  &  1290 & $20\times20$ & 1864  & 1775  &  1483 \\    
    02 & $20\times15$ & 1789  & 1585  &  1386 & $20\times20$ & 2116  & 1969  &  1669 \\
    03 & $20\times15$ & 1868  & 1660  &  1424 & $20\times20$ & 2225  & 1963  &  1697 \\
    04 & $20\times15$ & 1896  & 1714  &  1437 & $20\times20$ & 1989  & 1813  &  1510 \\
    05 & $20\times15$ & 1767  & 1606  &  1383 & $20\times20$ & 1749  & 1825  &  1539 \\
    06 & $20\times15$ & 1564  & 1496  &  1264 & $20\times20$ & 1950  & 1817  &  1524 \\
    07 & $20\times15$ & 1603  & 1467  &  1315 & $20\times20$ & 2065  & 1820  &  1614 \\
    08 & $20\times15$ & 1652  & 1648  &  1316 & $20\times20$ & 2103  & 1718  &  1512 \\
    09 & $20\times15$ & 1611  & 1576  &  1319 & $20\times20$ & 1914  & 1778  &  1482 \\
    10 & $20\times15$ & 1660  & 1605  &  1335 & $20\times20$ & 2001  & 1947  &  1597 \\
    \hline    
    01 & $30\times15$ & 2154  & 1949  &  1691 & $30\times20$ & 2390  & 2136  &  1939 \\    
    02 & $30\times15$ & 2473  & 2108  &  1822 & $30\times20$ & 2501  & 2398  &  2056 \\
    03 & $30\times15$ & 2430  & 2129  &  2078 & $30\times20$ & 2269  & 2097  &  1835  \\
    04 & $30\times15$ & 2099  & 1922  &  1600 & $30\times20$ & 2470  & 2246  &  1936 \\
    05 & $30\times15$ & 2178  & 2038  &  1828 & $30\times20$ & 2420  & 2186  &  1958 \\
    06 & $30\times15$ & 2158  & 1955  &  1778 & $30\times20$ & 2414  & 2081  &  1824  \\
    07 & $30\times15$ & 2193  & 1993  &  1863 & $30\times20$ & 2639  & 2273  &  1921 \\
    08 & $30\times15$ & 2219  & 1996  &  1801 & $30\times20$ & 2233  & 2225  &  1804 \\
    09 & $30\times15$ & 2354  & 1950  &  1750 & $30\times20$ & 2569  & 2350  &  1973 \\
    10 & $30\times15$ & 1938  & 1785  &  1550 & $30\times20$ & 2491  & 2312  &  2001 \\
    \hline     
    01 & $50\times15$ & 3238  & 2886  &  2781 & $50\times20$ & 3732  & 3164  &  3047 \\    
    02 & $50\times15$ & 3571  & 3119  &  2954 & $50\times20$ & 3477  & 3151  &  2760 \\
    03 & $50\times15$ & 3114  & 2848  &  2787 & $50\times20$ & 3381  & 3183  &  2969 \\
    04 & $50\times15$ & 3111  & 2859  &  2845 & $50\times20$ & 3446  & 3022  &  2847 \\
    05 & $50\times15$ & 3074  & \textbf{2924}  &  2924 & $50\times20$ & 3267  & 2991  &  2775 \\
    06 & $50\times15$ & 3113  & 2801  &  2588 & $50\times20$ & 3565  & 3250  &  2962 \\
    07 & $50\times15$ & 3230  & \textbf{3036}  &  3036 & $50\times20$ & 3678  & 3157  &  2920 \\
    08 & $50\times15$ & 3448  & 3030  &  2969 & $50\times20$ & 3723  & 3322  &  2919 \\
    09 & $50\times15$ & 3437  & 3164  &  2810 & $50\times20$ & 3538  & 3197  &  2873 \\
    10 & $50\times15$ & 3397  & 2879  &  2873 & $50\times20$ & 3456  & \textbf{3241}  &  3241  \\
    \hline      
    01 & $100\times15$ & 5945  & \textbf{5493}  &  5493 & $100\times20$ & 6215  & \textbf{5505}  &  5505 \\    
    02 & $100\times15$ & 5683  & \textbf{5517}  &  5517 & $100\times20$ & 5978  & \textbf{5663}  &  5663 \\
    03 & $100\times15$ & 5457  & \textbf{5212}  &  5212 & $100\times20$ & 6136  & 5465  &  5443 \\
    04 & $100\times15$ & 6088  & \textbf{5802}  &  5802 & $100\times20$ & 6193  & 5593  &  5511 \\
    05 & $100\times15$ & 5701  & \textbf{5568}  &  5568 & $100\times20$ & 6147  & \textbf{6066}  &  6066 \\
    06 & $100\times15$ & 6094  & \textbf{5821}  &  5821 & $100\times20$ & 6176  & \textbf{5699}  &  5699 \\
    07 & $100\times15$ & 6019  & \textbf{5756}  &  5756 & $100\times20$ & 6054  & 5858  &  5574\\
    08 & $100\times15$ & 5898  & \textbf{5876}  &  5876 & $100\times20$ & 6040  & 5564  &  5505\\
    09 & $100\times15$ & 5545  & \textbf{5367}  &  5367 & $100\times20$ & 5981  & \textbf{5886}  &  5886 \\
    10 & $100\times15$ & 5784  & \textbf{5303}  &  5303 & $100\times20$ & 6089  & \textbf{5689}  &  5689 \\
    \hline    
\end{tabular}
\end{table*}

\begin{table*}[htb!]
    \caption{Part 2 of makespans of all JSP instances of Table \ref{tab:app_details_Large_dataset_makespan}.}
    \label{tab:app_Large_dataset_makespan2}
    \centering
\begin{tabular}{c|c|c|c|c||c|c|c|c}
    \hline
    Instance  & $n \times m$ & L2D & RS & OR-Tools & $n \times m$ & L2D & RS & OR-Tools\\
    \hline
    \hline
    01 & $150\times15$ & 8902  & \textbf{8470}  &  8470 & $150\times20$ & 8630  & \textbf{8146}  &  8146 \\    
    02 & $150\times15$ & 8077  & \textbf{7871}  &  7871 & $150\times20$ & 8258  & \textbf{8139}  &  8139 \\
    03 & $150\times15$ & 8281  & \textbf{7964}  &  7964 & $150\times20$ & 8693  & \textbf{8178}  &  8178 \\
    04 & $150\times15$ & 8268  & \textbf{8215}  &  8215 & $150\times20$ & 8717  & \textbf{8219}  &  8219 \\
    05 & $150\times15$ & 8423  & \textbf{7974}  &  7974 & $150\times20$ & 8720  & \textbf{8663}  &  8663\\
    06 & $150\times15$ & 8544  & \textbf{8506}  &  8506 & $150\times20$ & 8577  & \textbf{8086}  &  8086 \\
    07 & $150\times15$ & 8703  & \textbf{8227}  &  8227 & $150\times20$ & 8622  & 7966  &  7962 \\
    08 & $150\times15$ & 8265  & \textbf{7940}  &  7940 & $150\times20$ & 8395  & \textbf{8101}  &  8101 \\
    09 & $150\times15$ & 8522  & \textbf{8472}  &  8472 & $150\times20$ & 8308  & \textbf{8000}  &  8000 \\
    10 & $150\times15$ & 8099  & \textbf{7937}  &  7937 & $150\times20$ & 8576  & \textbf{8215}  &  8215 \\
    \hline    
    01 & $200\times15$ & 11018  & \textbf{10696}  &  10696 & $200\times20$ & 10687  & \textbf{10545}  &  10545 \\    
    02 & $200\times15$ & 10770  & \textbf{10576}  &  10576 & $200\times20$ & 11453  & \textbf{10464}  &  10464\\
    03 & $200\times15$ & 11030  & \textbf{10952}  &  10952 & $200\times20$ & 12069  & \textbf{11040}  &  11040\\
    04 & $200\times15$ & \textbf{10881}  & \textbf{10881}  &  10881 & $200\times20$ & 11044  & \textbf{10644}  &  10644 \\
    05 & $200\times15$ & 10706  & \textbf{10701}  &  10701 & $200\times20$ & 11070  & \textbf{10793}  &  10793 \\
    06 & $200\times15$ & 10540  & \textbf{10510}  &  10510 & $200\times20$ & 11029  & \textbf{10427}  &  10427 \\
    07 & $200\times15$ & 10810  & \textbf{10345}  &  10345 & $200\times20$ & 10972  & \textbf{10699}  &  10699 \\
    08 & $200\times15$ & \textbf{10355}  & \textbf{10355}  &  10355 & $200\times20$ & 11033  & \textbf{10710}  &  10710 \\
    09 & $200\times15$ & 10983  & \textbf{10668}  &  10668 & $200\times20$ & 10988  & \textbf{10811}  &  10811 \\
    10 & $200\times15$ & 11036  & \textbf{10969}  &  10969 & $200\times20$ & 10787  & \textbf{10666}  &  10666 \\
    \hline
\end{tabular}
\end{table*}

\begin{table*}[htb!]
    \caption{Makespans of all MK instances (FJSP).}
\label{tab:app_FJSP_MK_datasetDetails}
    \centering
\begin{tabular}{c|c|cc|c}
    \hline
    Instance  & $n \times m$ & RS & RS+100 & OPT \\
    \hline
    \hline
    mk01 & $10\times6$  &  42    &  41 & 39 \\
    mk02 & $10\times6$  &  36    &  35 & 26 \\
    mk03 & $15\times8$  & \textbf{204}    & \textbf{204} & 204 \\
    mk04 & $15\times8$  &  75    &  67 & 60 \\
    mk05 & $15\times4$  & 185    & 180 & 172 \\
    mk06 & $10\times15$ &  97   &  81 & 58 \\
    mk07 & $20\times5$  & 216    & 188 & 139 \\
    mk08 & $20\times10$ & \textbf{523}   & \textbf{523} & 523 \\
    mk09 & $20\times10$ & 316   & 311 & 307 \\
    mk10 & $20\times15$ & 252   & 240 & 197 \\
    \hline
\end{tabular}
\end{table*}

\begin{table*}[htb!]
    \caption{Makespans of all LA instances (FJSP).}
\label{tab:app_FJSP_LA_datasetDetails}
    \centering
\begin{tabular}{c|c|ccc|ccc|ccc}
    \hline
     &  & \multicolumn{3}{c|}{edata} & \multicolumn{3}{c|}{rdata} &  \multicolumn{3}{c}{vdata} \\
    \hline
    Instance  & $n \times m$ & RS & RS+100 & OPT & RS & RS+100 & OPT & RS & RS+100 & OPT\\
    \hline
    \hline
    la01 & $10\times 5$   &  721 &  621 &  609 &  595 &  586 &  571 &  587 &  575 & 570 \\
    la02 & $10\times 5$   &  799 &  747 &  655 &  567 &  549 &  530 &  553 &  535 & 529 \\
    la03 & $10\times 5$   &  628 &  586 &  550 &  520 &  489 &  478 &  486 &  484 & 477 \\
    la04 & $10\times 5$   &  685 &  623 &  568 &  567 &  513 &  502 &  546 &  513 & 502 \\
    la05 & $10\times 5$   &  593 &  517 &  503 &  483 &  464 &  457 &  520 &  461 & 457 \\
    \hline
    la06 & $15\times 5$   &  847 &  \textbf{833} &  833 &  822 &  802 &  799 &  818 &  806 & 799 \\
    la07 & $15\times 5$   &  931 &  809 &  762 &  820 &  751 &  750 &  776 &  754 & 749 \\
    la08 & $15\times 5$   &  940 &  860 &  845 &  791 &  769 &  765 &  774 &  771 & 765 \\
    la09 & $15\times 5$   &  949 &  900 &  878 &  888 &  861 &  853 &  875 &  860 & 853 \\
    la10 & $15\times 5$   &  879 &  \textbf{866} &  866 &  845 &  816 &  804 &  821 &  808 & 804 \\
    \hline
    la11 & $20\times 5$   & 1208 & 1107 & 1103 & 1081 & 1074 & 1071 & 1082 & 1073 &1071 \\
    la12 & $20\times 5$   & 1001 &  979 &  960 &  959 &  938 &  936 &  946 &  938 & 936 \\
    la13 & $20\times 5$   & 1150 & 1074 & 1053 & 1048 & 1041 & 1038 & 1040 & 1040 &1038 \\
    la14 & $20\times 5$   & 1195 & \textbf{1123} & 1123 & 1079 & 1074 & 1070 & 1077 & 1071 &1070 \\
    la15 & $20\times 5$   & 1390 & 1269 & 1111 & 1176 & 1101 & 1090 & 1098 & 1091 &1089 \\
    \hline
    la16 & $10\times 10$  & 1002 &  952 &  892 &  870 &  751 &  717 &  763 &  \textbf{717} & 717 \\
    la17 & $10\times 10$  &  794 &  781 &  707 &  767 &  682 &  646 &  690 &  \textbf{646} & 646 \\
    la18 & $10\times 10$  &  911 &  897 &  842 &  751 &  717 &  666 &  \textbf{663} &  \textbf{663} & 663 \\
    la19 & $10\times 10$  &  973 &  915 &  796 &  860 &  787 &  700 &  652 &  628 & 617 \\
    la20 & $10\times 10$  & 1043 &  935 &  857 &  955 &  809 &  756 &  767 &  \textbf{756} & 756 \\
    \hline
    la21 & $15\times 10$  & 1248 & 1136 & 1017 &  956 &  916 &  835 &  820 &  817 & 806 \\
    la22 & $15\times 10$  & 993  &  980 &  882 &  888 &  850 &  760 &  784 &  754 & 739 \\
    la23 & $15\times 10$  & 1069 & 1038 &  950 &  918 &  897 &  842 &  836 &  821 & 815 \\
    la24 & $15\times 10$  & 1045 & 1025 &  909 &  908 &  883 &  808 &  812 &  794 & 777 \\
    la25 & $15\times 10$  & 1026 & 1016 &  941 &  908 &  866 &  791 &  876 &  771 & 756 \\
    \hline
    la26 & $20\times 10$  & 1344 & 1239 & 1125 & 1188 & 1112 & 1061 & 1068 & 1059 &1054 \\
    la27 & $20\times 10$  & 1419 & 1347 & 1186 & 1192 & 1152 & 1091 & 1116 & 1095 &1085 \\
    la28 & $20\times 10$  & 1402 & 1294 & 1149 & 1167 & 1124 & 1080 & 1100 & 1077 &1070 \\
    la29 & $20\times 10$  & 1325 & 1293 & 1118 & 1074 & 1038 &  998 & 1014 & 1001 & 994 \\
    la30 & $20\times 10$  & 1433 & 1373 & 1204 & 1185 & 1135 & 1078 & 1080 & 1078 &1069 \\
    \hline
    la31 & $30\times 10$  & 1748 & 1692 & 1539 & 1573 & 1541 & 1521 & 1544 & 1525 &1520 \\
    la32 & $30\times 10$  & 1896 & 1819 & 1698 & 1711 & 1698 & 1659 & 1673 & 1665 &1658 \\
    la33 & $30\times 10$  & 1716 & 1697 & 1547 & 1580 & 1512 & 1499 & 1514 & 1503 &1497 \\
    la34 & $30\times 10$  & 1901 & 1723 & 1604 & 1582 & 1552 & 1536 & 1545 & 1542 &1535 \\
    la35 & $30\times 10$  & 1911 & 1817 & 1736 & 1621 & 1583 & 1550 & 1564 & 1558 &1549 \\
    \hline
    la36 & $15\times 15$  & 1298 & 1254 & 1162 & 1146 & 1138 & 1030 &  965 &  955 & 948 \\
    la37 & $15\times 15$  & 1592 & 1498 & 1397 & 1239 & 1204 & 1077 & 1026 &  993 & 986 \\
    la38 & $15\times 15$  & 1407 & 1302 & 1144 & 1195 & 1083 &  962 &  \textbf{943} &  \textbf{943} & 943 \\
    la39 & $15\times 15$  & 1421 & 1275 & 1184 & 1243 & 1118 & 1024 &  984 &  935 & 922 \\
    la40 & $15\times 15$  & 1321 & 1291 & 1150 & 1104 & 1080 &  970 &  961 &  \textbf{955} & 955 \\
    \hline
\end{tabular}
\end{table*}

\begin{table*}[htb!]
    \caption{Part 1 of makespans of all TA instances (JSP).}
\label{tab:app_TAdataset_makespan}
    \centering
\begin{tabular}{c|c|ccc|ccc|c|c}
    \hline
    Instance  & $n \times m$ & SPT & FIFO & MOR & Park & L2D & SchN & RS & OPT \\
    \hline
    \hline
    Ta01 & $15\times15$ & 1462 & 1486 & 1438 & 1389 & 1443 & 1452 & \textbf{1356} & 1231 \\
    Ta02 & $15\times15$ & 1446 & 1486 & 1452 & 1519 & 1544 & 1411 & \textbf{1372} & 1244 \\
    Ta03 & $15\times15$ & 1495 & 1461 & 1418 & 1457 & 1440 & \textbf{1396} & 1467 & 1218 \\
    Ta04 & $15\times15$ & 1708 & 1575 & 1457 & 1465 & 1637 & \textbf{1348} & 1424 & 1175 \\
    Ta05 & $15\times15$ & 1618 & 1457 & 1448 & \textbf{1352} & 1619 & 1382 & 1378 & 1224 \\
    Ta06 & $15\times15$ & 1522 & 1528 & 1486 & 1481 & 1601 & 1413 & \textbf{1382} & 1238 \\
    Ta07 & $15\times15$ & 1434 & 1497 & 1456 & 1554 & 1568 & 1380 & \textbf{1356} & 1227 \\
    Ta08 & $15\times15$ & 1457 & 1496 & 1482 & 1488 & 1468 & \textbf{1374} & 1438 & 1217 \\
    Ta09 & $15\times15$ & 1622 & 1642 & 1594 & 1556 & 1627 & 1523 & \textbf{1497} & 1274 \\
    Ta10 & $15\times15$ & 1697 & 1600 & 1582 & 1501 & 1527 & 1493 & \textbf{1442} & 1241 \\
    \hline
    Ta11 & $20\times15$ & 1865 & 1701 & 1665 & 1626 & 1794 & 1612 & \textbf{1590} & 1357 \\
    Ta12 & $20\times15$ & 1667 & 1670 & 1739 & 1668 & 1805 & 1600 & \textbf{1574} & 1367 \\
    Ta13 & $20\times15$ & 1802 & 1862 & 1642 & 1715 & 1932 & 1625 & \textbf{1555} & 1342 \\
    Ta14 & $20\times15$ & 1635 & 1812 & 1662 & 1642 & 1664 & 1590 & \textbf{1511} & 1345 \\
    Ta15 & $20\times15$ & 1835 & 1788 & 1682 & 1672 & 1730 & 1676 & \textbf{1557} & 1339 \\
    Ta16 & $20\times15$ & 1965 & 1825 & 1638 & 1700 & 1710 & 1550 & \textbf{1549} & 1360 \\
    Ta17 & $20\times15$ & 2059 & 1899 & 1856 & \textbf{1678} & 1897 & 1753 & 1739 & 1462 \\
    Ta18 & $20\times15$ & 1808 & 1833 & 1710 & 1684 & 1794 & \textbf{1668} & 1705 & 1396 \\
    Ta19 & $20\times15$ & 1789 & 1716 & 1651 & 1900 & 1682 & 1622 & \textbf{1571} & 1332 \\
    Ta20 & $20\times15$ & 1710 & 1827 & 1622 & 1752 & 1739 & 1604 & \textbf{1560} & 1348 \\
    \hline
    Ta21 & $20\times20$ & 2175 & 2089 & 1964 & 2199 & 2252 & 1921 & \textbf{1910} & 1642 \\
    Ta22 & $20\times20$ & 1965 & 2146 & 1905 & 2049 & 2102 & 1844 & \textbf{1758} & 1600 \\
    Ta23 & $20\times20$ & 1933 & 2010 & 1922 & 2006 & 2085 & 1879 & \textbf{1867} & 1557 \\
    Ta24 & $20\times20$ & 2230 & 1989 & 1943 & 2020 & 2200 & 1922 & \textbf{1912} & 1644 \\
    Ta25 & $20\times20$ & 1950 & 2160 & 1957 & 1981 & 2201 & \textbf{1897} & 1952 & 1595 \\
    Ta26 & $20\times20$ & 2188 & 2182 & 1964 & 2057 & 2176 & \textbf{1887} & 1918 & 1643 \\
    Ta27 & $20\times20$ & 2096 & 2091 & 2160 & 2187 & 2132 & \textbf{2009} & 2015 & 1680 \\
    Ta28 & $20\times20$ & 1968 & 1980 & 1952 & 2054 & 2146 & \textbf{1813} & 1866 & 1603 \\
    Ta29 & $20\times20$ & 2166 & 2011 & 1899 & 2210 & 1952 & \textbf{1875} & 1916 & 1625 \\
    Ta30 & $20\times20$ & 1999 & 1941 & 2017 & 2140 & 2035 & 1913 & \textbf{1804} & 1584 \\
    \hline
    Ta31 & $30\times15$ & 2335 & 2277 & 2143 & 2251 & 2565 & 2055 & \textbf{2004} & 1764 \\
    Ta32 & $30\times15$ & 2432 & 2279 & 2188 & 2378 & 2388 & 2268 & \textbf{2097} & 1784 \\
    Ta33 & $30\times15$ & 2453 & 2481 & 2308 & 2316 & 2324 & 2281 & \textbf{2059} & 1791 \\
    Ta34 & $30\times15$ & 2434 & 2546 & 2193 & 2319 & 2332 & \textbf{2061} & 2093 & 1829 \\
    Ta35 & $30\times15$ & 2497 & 2478 & 2255 & 2333 & 2505 & 2218 & \textbf{2179} & 2007 \\
    Ta36 & $30\times15$ & 2445 & 2433 & 2307 & 2210 & 2497 & 2154 & \textbf{2102} & 1819 \\
    Ta37 & $30\times15$ & 2664 & 2382 & 2190 & 2201 & 2325 & 2112 & \textbf{1968} & 1771 \\
    Ta38 & $30\times15$ & 2155 & 2277 & 2179 & 2151 & 2302 & \textbf{1970} & 1999 & 1673 \\
    Ta39 & $30\times15$ & 2477 & 2255 & 2167 & 2138 & 2410 & 2146 & \textbf{1978} & 1795 \\
    Ta40 & $30\times15$ & 2301 & 2069 & 2028 & 2007 & 2140 & 2030 & \textbf{1986} & 1669 \\
    \hline
    Ta41 & $30\times20$ & 2499 & 2543 & 2538 & 2654 & 2667 & 2572 & \textbf{2435} & 2005 \\
    Ta42 & $30\times20$ & 2710 & 2669 & 2440 & 2579 & 2664 & 2397 & \textbf{2253} & 1937 \\
    Ta43 & $30\times20$ & 2434 & 2506 & 2432 & 2737 & 2431 & 2310 & \textbf{2190} & 1846 \\
    Ta44 & $30\times20$ & 2906 & 2540 & 2426 & 2772 & 2714 & 2456 & \textbf{2343} & 1979 \\
    Ta45 & $30\times20$ & 2640 & 2565 & 2487 & 2435 & 2637 & 2445 & \textbf{2337} & 2000 \\
    Ta46 & $30\times20$ & 2667 & 2582 & 2490 & 2681 & 2776 & 2541 & \textbf{2330} & 2004 \\
    Ta47 & $30\times20$ & 2620 & 2508 & 2286 & 2428 & 2476 & 2280 & \textbf{2270} & 1889 \\
    Ta48 & $30\times20$ & 2620 & 2541 & 2371 & 2440 & 2490 & 2358 & \textbf{2279} & 1941 \\
    Ta49 & $30\times20$ & 2666 & 2550 & 2397 & 2446 & 2556 & 2301 & \textbf{2249} & 1961 \\
    Ta50 & $30\times20$ & 2429 & 2531 & 2469 & 2530 & 2628 & 2453 & \textbf{2251} & 1923 \\
    \hline
\end{tabular}
\end{table*}

\begin{table*}[htb!]
    \caption{Part 2 of makespans of all TA instances (JSP).}
\label{tab:app_TAdataset_makespan2}
    \centering
\begin{tabular}{c|c|ccc|ccc|c|c}
    \hline
    Instance  & $n \times m$ & SPT & FIFO & MOR & Park & L2D & SchN & RS & OPT \\
    \hline
    \hline
    Ta51 & $50\times15$ &  3856 & 3590 & 3567 & 3145 & 3599 & 3382 &  \textbf{2989} & 2760 \\
    Ta52 & $50\times15$ &  3266 & 3365 & 3303 & 3157 & 3341 & 3231 &  \textbf{2986} & 2756 \\
    Ta53 & $50\times15$ &  3507 & 3169 & 3115 & 3103 & 3186 & 3083 &  \textbf{2851} & 2717 \\
    Ta54 & $50\times15$ &  3142 & 3218 & 3265 & 3278 & 3266 & 3068 &  \textbf{2913} & 2839 \\
    Ta55 & $50\times15$ &  3225 & 3291 & 3279 & 3142 & 3232 & 3078 &  \textbf{2915} & 2679 \\
    Ta56 & $50\times15$ &  3530 & 3329 & 3100 & 3258 & 3378 & 3065 &  \textbf{2982} & 2781 \\
    Ta57 & $50\times15$ &  3725 & 3654 & 3335 & 3230 & 3471 & 3266 &  \textbf{3126} & 2943 \\
    Ta58 & $50\times15$ &  3365 & 3362 & 3420 & 3469 & 3732 & 3321 &  \textbf{3108} & 2885 \\
    Ta59 & $50\times15$ &  3294 & 3357 & 3117 & 3108 & 3381 & 3044 &  \textbf{2887} & 2655 \\
    Ta60 & $50\times15$ &  3500 & 3129 & 3044 & 3256 & 3352 & 3036 &  \textbf{2861} & 2723 \\
    \hline
    Ta61 & $50\times20$ &  3606 & 3690 & 3376 & 3425 & 3654 & 3202 &  \textbf{3133} & 2868 \\
    Ta62 & $50\times20$ &  3639 & 3657 & 3417 & 3626 & 3722 & 3339 &  \textbf{3234} & 2869 \\
    Ta63 & $50\times20$ &  3521 & 3367 & 3276 & \textbf{3110} & 3536 & 3118 & 3130 & 2755 \\
    Ta64 & $50\times20$ &  3447 & 3179 & 3057 & 3329 & 3631 & 2989 &  \textbf{2982} & 2702 \\
    Ta65 & $50\times20$ &  3332 & 3273 & 3249 & 3339 & 3359 & 3168 &  \textbf{3007} & 2725 \\
    Ta66 & $50\times20$ &  3677 & 3610 & 3335 & 3340 & 3555 & 3199 &  \textbf{3153} & 2845 \\
    Ta67 & $50\times20$ &  3487 & 3612 & 3392 & 3371 & 3567 & 3236 &  \textbf{3054} & 2825 \\
    Ta68 & $50\times20$ &  3336 & 3471 & 3251 & 3265 & 3680 & 3072 &  \textbf{2991} & 2784 \\
    Ta69 & $50\times20$ &  3862 & 3607 & 3526 & 3798 & 3592 & 3535 &  \textbf{3305} & 3071 \\
    Ta70 & $50\times20$ &  3801 & 3784 & 3590 & 3919 & 3643 & 3436 &  \textbf{3304} & 2995 \\
    \hline
    Ta71 & $100\times20$ & 6232 & 6270 & 5938 & 5962 & 6452 & 5879 & \textbf{5679} & 5464 \\
    Ta72 & $100\times20$ & 5973 & 5671 & 5639 & 5522 & 5695 & 5456 & \textbf{5297} & 5181 \\
    Ta73 & $100\times20$ & 6482 & 6357 & 6128 & 6335 & 6462 & 6052 & \textbf{5807} & 5568 \\
    Ta74 & $100\times20$ & 6062 & 6003 & 5642 & 5827 & 5885 & 5513 & \textbf{5413} & 5339 \\
    Ta75 & $100\times20$ & 6217 & 6420 & 6212 & 6042 & 6355 & 5992 & \textbf{5623} & 5392 \\
    Ta76 & $100\times20$ & 6370 & 6183 & 5936 & 5707 & 6135 & 5773 & \textbf{5487} & 5342 \\
    Ta77 & $100\times20$ & 6045 & 5952 & 5829 & 5737 & 6056 & 5637 & \textbf{5475} & 5436 \\
    Ta78 & $100\times20$ & 6143 & 6328 & 5886 & 5979 & 6101 & 5833 & \textbf{5426} & 5394 \\
    Ta79 & $100\times20$ & 6018 & 6003 & 5652 & 5799 & 5943 & 5556 & \textbf{5411} & 5358 \\
    Ta80 & $100\times20$ & 5848 & 5763 & 5707 & 5718 & 5892 & 5545 & \textbf{5448} & 5183 \\
    \hline
\end{tabular}
\end{table*}

\end{document}